\pdfoutput=1
\documentclass[onecolumn]{article}
\usepackage{style_arxiv}
\usepackage{amsmath,amsfonts,bm}

\def\eqref#1{equation~\ref{#1}}

\def\1{\bm{1}}

\DeclareMathAlphabet{\mathsfit}{\encodingdefault}{\sfdefault}{m}{sl}
\SetMathAlphabet{\mathsfit}{bold}{\encodingdefault}{\sfdefault}{bx}{n}

\theoremstyle{plain}
\newtheorem{result}{Result}

\theoremstyle{definition}

\theoremstyle{remark}

\theoremstyle{plain}
\newtheorem*{resultr}{Result}
\newtheorem*{resultbis}{Result \theresult\ bis}

\newcommand{\tmax}[0]{t_{\rm max}}
\newlength{\figwidth}
\newcommand{\figscale}{1.}
\AtBeginDocument{\setlength{\figwidth}{\figscale\textwidth}}

\newcommand{\vx}{\bm{\mathrm{x}}}
\newcommand{\vs}{\bm{\mathrm{s}}}
\newcommand{\vb}{\bm{\mathrm{b}}}
\newcommand{\vz}{\bm{\mathrm{z}}}
\newcommand{\vw}{\bm{\mathrm{w}}}
\newcommand{\vu}{\bm{\mathrm{u}}}
\newcommand{\vv}{\bm{\mathrm{v}}}
\newcommand{\vM}{\bm{\mathrm{M}}}
\newcommand{\vxi}{\bm{\xi}}
\newcommand{\vtheta}{\bm{\theta}}
\newcommand{\vmu}{\bm{\mu}}
\newcommand{\vI}{\bm{I}}
\newcommand{\dd}{\mathrm{d}}
\newcommand{\sign}{\mathrm{sign}}

\title{Weighting Schedules Govern What and When Score-Based Generative Models Learn from Multimodal Data}

\author[1]{J\'er\'emie Klinger%
\thanks{Corresponding author: \href{mailto:jeremie.klinger@phys.ens.psl.eu}{\texttt{jeremie.klinger@phys.ens.psl.eu}}}%
}
\author[1]{Rapha\"el Urfin}
\author[1]{Giulio Biroli}
\author[1]{Marylou Gabri\'e}
\affil[1]{Laboratoire de Physique de l'\'Ecole normale sup\'erieure, ENS, Universit\'e PSL, CNRS, Sorbonne Universit\'e, Universit\'e Paris Cit\'e, F-75005 Paris, France}
\date{\vspace{-7ex}}

\begin{document}

\pagenumbering{arabic}
\maketitle
\selectlanguage{american}

\begin{abstract}
    Score-based generative models generate new samples by integrating a time-dependent drift that carries Gaussian noise onto the target distribution. In practice this drift is modeled by a neural network, trained on a loss integrated over time $t$ with a weighting schedule $w(t)$. Along the backward dynamics, and for multi-modal distributions,  trajectories commit to modes of the target within a narrow  time window, the \textit{speciation time}. 
  In this work, focusing on high-dimensional
data, we decompose the integrated loss into its single-time contributions and
analyze each at fixed signal-to-noise ratio $\Lambda(t)$: we show that
$\Lambda(t)$ sets the rate at which each feature of a multimodal
target---the mode directions and their relative weights---is acquired during
training. Crucially, at high $\Lambda(t)$ all mode directions are acquired together, on a single timescale insensitive to their amplitudes, while the relative weights are not learned at all. Only near the speciation time, where $\Lambda(t)$ becomes of order one, do all features become learnable, each on its own timescale: the weights are acquired jointly with the directions, and the directions at rates set by their relative amplitudes. For models trained on time-integrated objectives, the learning dynamics is then governed by how much of the weighting effectively sits near the speciation time, which provides insights on $w(t)$ design choices.  These results follow from an exact high-dimensional analysis of the training dynamics of unbalanced and hierarchical Gaussian mixtures. 
Numerical experiments on image and human genome haplotype generation recover the predicted hierarchy of learning timescales in more complex settings. \looseness=-1

\end{abstract}

\vspace{1cm}

\vspace{-5ex}

\section{Introduction}
Over the last decade, the fast-paced development of transport-based generative models has led to tremendous progress in diverse and complex sampling tasks across both natural and artificial modalities.
Vision modeling represents one such novelty, with latent diffusion models constituting state-of-the-art image \citep{dhariwal2021diffusionmodelsbeatgans, Rombach_2022} and video generators \citep{sora2024, yang2025cogvideox}.
In the physical and chemical sciences, Boltzmann generators \citep{noe2019boltzmann} offer an alternative to traditional 
sampling methods such as MCMC, either directly replacing them 
or augmenting them \citep{gabrie2022adaptive, grenioux2025amorphous,rehman2026autoregressive}.
While this empirical progress 
has repeatedly outpaced our ability to explain it, 
significant leverage still lies in a precise theoretical understanding of training and the subsequent generative process.

The unified framework of Stochastic Interpolants \citep[SI,][]{albergo2023stochastic, Albergo2025unifying} proposes a shared theoretical setting to discuss score-based generative models. Building such models amounts to optimizing a drift field driving Gaussian samples towards the target distribution, with generation quality directly impacted by practitioners' {\bf training} and {\bf sampling} choices \citep{karras2022elucidatingdesignspacediffusionbased}.
Within this paradigm, the statistical physics analysis of reverse time dynamics \citep{biroli_2024} underlying the generation process has led to the identification of a narrow critical timescale dubbed speciation time: the brief window in which the generative trajectory commits to one mode of the target distribution, plainly identifying where compute effort should be spent during the {\bf sampling phase.}
While the practical impact of this result is undeniable, it assumes the optimal drift to be known. In practice that drift is learned, and which features of the target the converged score actually resolves is decided by the choices made during the {\bf training phase}.

Chief among those is the weighting schedule, which sets the noise levels the objective emphasizes and drives much of the reported progress in sample quality and training efficiency \citep{choi2022perception, hang2023efficient, esser2024scaling}. These gains, however, rest largely on empirical tuning: how
the weighting schedule governs the acquisition of data features during training
still lacks a theoretical account. We bridge this gap by considering the training dynamics of two controlled families of structured data distributions: unbalanced and hierarchical Gaussian Mixture Models (GMM).
We show that the design of the training objective quantitatively sets the rate at which each feature — mode direction, relative weight, mode amplitude — is acquired.
In turn we empirically demonstrate that these results generalize to standard SI pipelines, with direct implications for the choice of weighting schedule. \looseness=-1

\paragraph*{Main contributions.}

Stochastic interpolant optimization is generically recast as minimizing

\begin{equation}
\label{eq:main_cont_obj}
    \mathcal{L}(\vtheta) = d^{-1}\mathbb{E}_{\vx_0, \vxi}\left[\int_0^{t_{\rm max}}\dd tw(t)\left\lVert \beta(t)\vs_{\vtheta}\left(\alpha(t)\vx_0 + \beta(t)\vxi,\, t\right) + \vxi\right\rVert^2\right] 
\end{equation}
with respect to the model parameters $\vtheta$, where $\vx_0$ is drawn from the target distribution $P_0$ and $\vxi$ from a standard Gaussian $\mathcal{N}(0, \vI_d)$. 
With unlimited training time and exact minimization, the integrated loss is
minimized by the exact score at every noise level, whatever the weighting
schedule $w$. In practice, however, training is stopped after a finite time and is 
at finite precision.
The noising schedule $(\alpha, \beta)$ and weighting schedule $w$ then constitute important design choices whose influence on learning dynamics for high-dimensional data is the primary concern of this paper. 

\begin{itemize}

    \item {\bf Training at fixed noise level -} 
    For both unbalanced and structured GMM variants characterized by finitely many modes with distinct directions and weights, we analyze the training dynamics of 
    a network $\vs_{\vtheta}$ specifically tailored to the GMM datasets at a single noising time $t^*$.
    The timescale on which each mode direction and mode weight are learned critically depends on $t^*$. In particular, for $t^*$ small with respect to the speciation time $t_s$, only the mode directions are recovered, on a timescale insensitive to the mode amplitudes, while focusing training around $t_s$ leads to weights and directions being learned jointly, at rates set by the modes' relative amplitude.

    \item {\bf Influence of the weighting schedule  -} 
    Using the signal-to-noise ratio as the shared noising metric, 
    we precisely show how the choice of weighting schedule $w$ imposes effective weights for the relevant noise levels.
    As a result, the dominant noising regions control the learning dynamics of mode direction and weights.
    Importantly, we emphasize that Diffusion and Flow Matching (FM)
    correspond to  different effective weights, resulting in distinct learning dynamics. \looseness=-1

    \item {\bf Empirical validation on real datasets -} We evaluate our results against 
    i) handmade unbalanced and hierarchical MNIST dataset and ii) Haplotypes from the Human Genome Dataset (HGD).
    We sweep over a breadth of SI frameworks, completing the GMM Denoising Score Matching experiments with Flow Matching, as well as the more recent Masked Discrete Diffusion framework for the HGD, 
    and recover quantitative agreement between theory and the empirical timescales governing hierarchical feature emergence.
    
\end{itemize}

\section*{Related works}

\paragraph*{Where structure appears at generation time.}
Mode commitment during the generation dynamics, first identified in \citet{Raya_2023}, is thoroughly analyzed in \citet{Biroli_2023, biroli_2024}: they introduce the speciation time, characterize it analytically for high-dimensional GMMs through overlap summary statistics, and generalize it to arbitrary distributions. The analysis is further extended to mixtures of strongly log-concave densities in \citet{li2024criticalwindowsnonasymptotictheory}.
\citet{Behjoo_2025, sclocchi2025phase} probe the same windows empirically through
forward-backward U-turn protocols on pretrained denoisers. All of these assume the score to be exact and locate a narrow
commitment window along the \emph{sampling} trajectory.
The complementary question, at what \emph{training}
time each feature becomes available, is the one we address.

\paragraph*{Where structure appears at training time.}
A series of works characterizes the optimal score reachable at finite sample complexity for a given neural network architecture, without studying the training dynamics: \citet{cui2024} consider a two-layer autoencoder at a fixed noising time for a symmetric bimodal GMM, completed by \citet{aranguri2025phaseaware}, who study mode weight recovery for asymmetric GMMs, while \citet{george2026denoisingscorematchingrandom} derive precise learning curves for random feature networks.
Another line of work focuses on the training dynamics itself. Outside the diffusion paradigm, \citet{Bachtis_2024} follow the training dynamics of energy-based models on hierarchically structured data, and find the levels of the hierarchy to be learned successively. Within the diffusion framework, \citet{cui2025} extend the two-layer autoencoder to all noising times, at the price of an encoding of time with low expressivity, \citet{nicoletti2026interplaydatastructureimbalance} track how data structure and imbalance shape the dynamics in a random feature model, and \citet{bardone2026theorylearningdatastatistics} show that moments of a distribution are learned sequentially.
Closest to our question, \citet{wang2026an} find the covariance eigendirections of a Gaussian target to display sequential alignment during training. Whereas all the structure lies in the covariance, we here focus on multimodal structure.

\paragraph*{Noising and weighting schedules.}
\citet{kingma2023understanding} show that most standard diffusion objectives are the
ELBO under different weightings; \citet{choi2022perception},
\citet{karras2022elucidatingdesignspacediffusionbased, hang2023efficient} and
\citet{esser2024scaling} propose state-of-the-art SNR-based schedules and weighting functions. On
the theory side, \citet{aranguri2025optimizingnoiseschedulesgenerative} optimize noise schedules
in high dimension, while \citet{aranguri2025phaseaware} analyze mode weight recovery using ad-hoc weighting schedules, and \citet{gagneux2025generationphases} study the influence of the choice of weighting scheme on the performance of trained neural networks. Instead, we here focus on learning of mode weights and hierarchy in  standard SI pipelines and analyze the consequences of the resulting weighting schemes.

\section{Setting}
\label{sec:setting}
\paragraph*{Stochastic interpolants.}
The Stochastic Interpolant (SI) framework \citep{albergo2023stochastic, Albergo2025unifying} encompasses both Diffusion Models (DM) \citep{sohl-dickstein_15,song2021scorebased, Ho_DDPM} and Flow Matching (FM) \citep{lipman2023flow}.
A SI defines a bridge between a target distribution $P_0$ and Gaussian white noise by introducing the interpolation
\begin{align}
\label{eq:stochastic_interpolants_0}
    \vx_t=\alpha(t)\vx_0+\beta(t)\vxi,
\end{align}
where $(\vx_0, \vxi)\sim P_0 \times  \mathcal{N}(0,\vI_d)$, $t\in[0,\tmax]$\footnote{For instance $\alpha(t)=e^{-t}, \ \beta(t)=\sqrt{1-e^{-2t}},\ \tmax=\infty$ for DM and $\alpha(t)=1-t, \ \beta(t)=t,\ \tmax=1$ for FM.}, and the schedule functions $(\alpha, \beta)$ defining the SI obey the boundary conditions $\alpha(0)=1,\ \beta(0)=0, \ \alpha(\tmax)=0,\ \beta(\tmax)=1$.
By construction, the marginal distribution $\vx_t \sim P_t$ interpolates between $P_0$ and $\mathcal{N}(0,\vI_d)$.
Following the pioneering works of \cite{Anderson_1982} and \cite{haussmann_1986}, the interpolant $\vx_t$ can also be written as the solution at time $t$ of the deterministic flow
\begin{align} 
\label{eq:ODE_flow}
\dot \vx(t)=\vb(\vx(t),t), ~~~\vx(t_{\rm max}) \sim \mathcal{N}(0,\vI_d)
\end{align}
integrated backward in time, with velocity field
\begin{align}
    \label{eq:gen_ode}
    \vb(x,t)=\mathbb{E}[\dot{\vx}_t\mid \vx_t=\vx]=\frac{\dot{\alpha}(t)}{\alpha(t)}\,\vx-\left(\dot{\beta}(t)-\frac{\dot{\alpha}(t)}{\alpha(t)}\beta(t)\right)\beta(t)\,\nabla_{\vx}\log P_t(\vx),
\end{align}
so that integrating \eqref{eq:ODE_flow} generates new samples from $P_0$. The term $\vs^*(\vx,t)\equiv\nabla_{\vx}\log P_t(\vx)$, dubbed the score function, is generally not available in closed form, but it is the minimizer of the Denoising Score Matching (DSM) loss \citep{hyvarinen_05,Vincent_2011}
\begin{align}
    \label{eq_DSM}
    \mathcal{L}_{\mathrm{DSM}}(\vs)=\tfrac 1 d \mathbb{E}_{(\vx_0, \vxi)\sim P_0\times \mathcal{N}(0,\vI_d)}\left[\int_0^{t_{\mathrm{max}}} \dd t w(t)\lVert \beta(t)\vs\left(\vx_t,t\right)+\vxi\rVert^2\right],
\end{align}
where the weighting function $w(t)$ is arbitrary.
In practice the score function is approximated by a neural network with parameters $\vtheta$ and optimized with local methods such as stochastic gradient descent \citep{robbins1951stochastic}.

Schedule and weighting functions $(\alpha,\beta,w)$ are crucial design choices: they set the relative weights of various noise levels and thus which features of the target distribution $P_0$ can possibly be recovered.
The precise interplay between these choices and the training or sampling dynamics remains poorly understood, and the few results available assume an exact, optimally trained score \citep{Biroli_2023, biroli_2024, aranguri2025optimizingnoiseschedulesgenerative}. They underlie our own analysis and are recalled below.

\paragraph*{Data distributions.}
We focus on two Gaussian mixtures with closed-form scores, one unbalanced and one hierarchical (see Fig.~\ref{fig:exscore-unbalanced} and  \ref{fig:exscore-kappa} for illustrative histograms):
\begin{itemize}
    \item $(\mathcal{D}_1)$, an unbalanced bimodal distribution,
\begin{equation}
\label{eq:unbalanced}
\begin{split}
    &P_0(\vx) = \omega\, \mathcal{N}(\vmu, \sigma^2 \vI_d) + (1-\omega)\, \mathcal{N}(-\vmu, \sigma^2 \vI_d),
    \quad ||\vmu||^2 = d, \quad \omega, \sigma^2 = O_d(1),\\
    &\nabla\log P_t(\vx)=-\frac{\vx}{\Gamma_t}+\frac{\alpha(t)\vmu}{\Gamma_t}\tanh\left(\frac{\alpha(t)\,\vmu^T\vx}{\Gamma_t} + \tfrac{1}{2}\log\tfrac{\omega}{1-\omega}\right),
    \end{split}
\end{equation}
    with $\Gamma_t=\alpha^2(t)\sigma^2+\beta^2(t)$.
    \item $(\mathcal{D}_2)$, a quadrimodal distribution inspired by \citet{Bachtis_2024},
\begin{equation}
\label{eq:hierarchical}
\begin{split}
    &P_0(\vx) = \frac{1}{4} \sum_{s_1 = \pm 1} \sum_{s_2 = \pm 1} \mathcal{N}(s_1 \vmu_1 + s_2 \vmu_2,\, \sigma^2 \vI_d),\\
    &\nabla\log P_t(\vx)=-\frac{\vx}{\Gamma_t}+\frac{\alpha(t)\vmu_1}{\Gamma_t}\tanh\left(\frac{\alpha(t)\,\vmu_1^T\vx}{\Gamma_t}\right)+\frac{\alpha(t)\vmu_2}{\Gamma_t}\tanh\left(\frac{\alpha(t)\,\vmu_2^T\vx}{\Gamma_t}\right),
\end{split}
\end{equation}
where $\vmu_1 \in \mathbb{R}^d$ is supported on the first $\kappa d$ coordinates and $\vmu_2$ on the remaining $(1-\kappa)d$ with $\lVert\vmu_1\rVert^2=\kappa d,\ \lVert\vmu_2\rVert^2=(1-\kappa )d,\ \kappa=O_d(1)$. We work in the high-dimensional limit $d\gg 1$, for two reasons: realistic data
are high-dimensional, and, as we show below, this is precisely the regime in
which the choice of $w(t)$ matters most.

\end{itemize}

\paragraph*{Diffusion Regimes and Speciation Time.}

Consider a symmetric bimodal target $P_0=\tfrac12\mathcal{N}(\vmu,\vI_d)+\tfrac12\mathcal{N}(-\vmu,\vI_d)$ and a diffusion model, i.e. $\alpha(t)=e^{-t},\ \beta(t)=\sqrt{1-e^{-2t}}$, $t\in[0,t_{\rm max}]$,
for which the exact score $\vs^*(\vx,t)$ is known and the reverse-time drift \eqref{eq:gen_ode} can be analyzed in closed form.
\citet{biroli_2024} show that the mode overlap\footnote{\S\ref{app:speciation} works with the rescaled overlap $\tilde q_t=\sqrt d\,q_t=\vmu^T\vx_t/\sqrt d$, there simply written $q$, which keeps the noise contribution $O_d(1)$ rather than $O_d(1/\sqrt d)$ and is the natural scale for the signal-vs-noise argument; the two differ only by this rescaling. \looseness=-1} $q_t=\vmu^T\vx_t/d$ evolves in an effective time-dependent potential $V_t(q)$, whose shape sharply shifts around the speciation time $t_s \equiv \frac{1}{2}\log(d)$.
In Regime~I, corresponding to times $t\gg t_s$, $V_t$ is effectively unimodal, and no information is retained about the bimodal structure of $P_0$.
Conversely, in Regime~II, $t\ll t_s$, $V_t$ consists of two well-separated harmonic wells centered on $\pm 1$: the bimodality of $P_0$ is fully resolved and the generated sample has committed to one of the modes.
The speciation time $t_s$ is the narrow window in which $V_t$ develops a cusp that decides the generation outcome, and specifies where sampling compute is best spent when integrating \eqref{eq:gen_ode}. The speciation phenomenon has been confirmed in other theoretical models \citep{Biroli_2023,ambrogioni2023statistical,li2024criticalwindowsnonasymptotictheory} and in real cases by numerical experiments \citep{Behjoo_2025,sclocchi2025phase}.

Note however that these results only concern sampling and they do not address how a trained score approaches $\vs^*$ as optimization of \eqref{eq_DSM} is carried out, nor whether $t_s$ plays a significant role, or whether the picture holds once modes carry structure beyond the simple bimodal balanced GMM - unequal mode weights, or a hierarchy of sub-modes at different scales. The aim of this paper is to study training dynamics under such structure, for general stochastic interpolants.

\section{Analytical results}

In the following, we study the training of a parametrized score model $\vs_{\vtheta}(\vx, t)$ by gradient descent on the population loss \eqref{eq_DSM}\footnote{We work throughout with the population loss i.e. the expectation value is exact. This departs from a line of work on the empirical loss, which asks how many samples are needed to learn a given feature \citep{bardone2026theorylearningdatastatistics} \looseness=-1}, in the infinitesimal learning rate limit known as Gradient Flow (GF)
\begin{align}
\label{eq:grad_flow}
    \frac{\dd\vtheta}{\dd\tau}=-\nabla_{\vtheta}\mathcal{L}_{\mathrm{DSM}}(\vs_{\vtheta}),
\end{align}
where the training time $\tau$ should not be confused with the sampling time $t$. 
The aim is to elucidate the dynamical timescale dependencies on the dimension $d$ and structure parameters $\omega$, $\kappa$.
To isolate dependencies between the schedules $(\alpha, \beta)$ and $w$, it is insightful to consider first a single noise level, equivalent to a Dirac weighting function $w(t) = \delta(t-t^*)$.  
In this case, the synthetic distributions $\mathcal{D}_{1,2}$, which allow for analytical treatment, explicitly showcase the sequential learning of structured data features.

\subsection{Single fixed noising time}
\label{sec:fixed_time}

\paragraph*{Score models.} At fixed noising time $t$, we parametrize the trainable score as a two-layer network with skip connection and $\tanh$ activation:
for $(\mathcal D_1)$,
\begin{align}
\label{eq:score_model_unbalanced}
    \vs_{\vtheta=(c,b,\vw)}(\vx,t)=c\vx+\vw\tanh(\vw^T\vx+b), \qquad c,b\in\mathbb R,\ \vw\in\mathbb R^d,
\end{align}
with the learned weight read out of the bias, $\rho=e^{2b}/(1+e^{2b})$; and for $(\mathcal D_2)$,
\begin{align}
\label{eq:score_model_4modes}
    \vs_{\vtheta=(c,\vw_1,\vw_2)}(\vx,t)=c\vx+\vw_1\tanh(\vw_1^T\vx)+\vw_2\tanh(\vw_2^T\vx),
\end{align}
where $\vw_1$ lives in the same block dimensional space as $\vmu_1$ (resp. for $(\vw_2, \vmu_2)$).
Under the GF dynamics \eqref{eq:grad_flow}, we show in appendix (\S\ref{app:unbalanced_exact}) and (\S\ref{app:hierarchical_exact}) that $\mathcal L_{\mathrm{DSM}}$ depends on $\vw$ (resp.\ $\vw_1,\vw_2$) only through its projection onto the mode direction(s) and its orthogonal component, so training reduces to a closed system of ODEs on the summary statistics
\begin{align}
\label{eq:order_params}
    m=\frac{\vw^T\vmu}{\alpha(t)\,d},\quad q=\frac{\lVert\vw^\perp\rVert^2}{\alpha(t)^2\,d}
    \qquad\left(\text{resp. } m_i=\frac{\vw_i^T\vmu_i}{\alpha(t)\,\kappa_id},\quad q_i=\frac{\lVert\vw_i^\perp\rVert^2}{\alpha(t)^2\,\kappa_id},\ i=1,2\right)
\end{align}
together with $c$ (and $b$ for $\mathcal D_1$), where $\vw^\perp$ is the component of $\vw$ orthogonal to $\vmu$. The parameterizations~\eqref{eq:score_model_unbalanced}
and~\eqref{eq:score_model_4modes} are expressive enough to represent the exact
score. Our aim is therefore to determine whether, and on what
timescales, the summary statistics reach their optimal values during training.

\paragraph*{Signal-to-noise ratio and speciation time for generic SI.} 
 
As shown in appendix (\S\ref{app:speciation}), the speciation time analysis carried out in \cite{biroli_2024} holds for generic SI schedules $(\alpha, \beta)$. Mode selection during sampling is now governed by the signal-to-noise (SNR) ratio
\begin{align}
\label{eq:snr_train}
    \Lambda(t)=\frac{\alpha^2(t)\,d}{\alpha^2(t)\sigma^2+\beta^2(t)},
\end{align}
redefining the speciation time by the convention $\Lambda(t_s)\equiv 1$ (equivalently $\alpha/\beta \sim d^{-1/2}$). 
The underlying principle is clear: mode mixing occurs at speciation and this analysis carries over to the training paradigm. Indeed, training deep in Regime~II, $\Lambda_t=O(d)$, exposes the network only to noise levels at which the data structure is fully resolved\footnote{This presupposes that the structural parameters $\omega$ themselves remain $O_d(1)$, contrary to the results presented in \citet{aranguri2025phaseaware}
 }; training around speciation exposes it to meaningfully collapsed data structures. 
By characterizing training at a given $\Lambda_t$ in Regime~II and around speciation, for each of $(\mathcal D_1), (\mathcal D_2)$, we show that its choice determines not only how fast $(m,q)$ are learned, but what is learned at all. Our findings are summarized in the following results.

\begin{result}[Training dynamics for $(\mathcal{D}_1)$ (\S\ref{app:unbalanced_exact})]
\label{thm:unbalanced}
Let $\vw(0)\in\mathbb S^{d-1}$ with $c(0),b(0)=O_d(1)$, and let $(c,\vw,b)$ evolve under GF \eqref{eq:grad_flow} for the score model \eqref{eq:score_model_unbalanced} at fixed $\Lambda_t$. In the $d\to\infty$ limit the $d$-dimensional flow closes on the summary statistics $(m,q,c,b)$, which obey a finite system of ODEs given in \S\ref{app:unbalanced_exact}. Specifically:
\begin{enumerate}[label=(\alph*)]
    \item \textbf{Regime~II} ($\Lambda_t=O_d(d)$). On $\tau=O_d(1)$ times $c\to -1/(\Gamma_t+\alpha_t^2)$, while $(m,q,b)$ stay at initialization. On $\tau=O_d(d)$ times 
    $(m,q,b)$ relax to $(1/\Gamma_t,0,b(0))$ while     
    $c$ relaxes instantaneously to $c^*(m,q)$.
    \item \textbf{Speciation} ($\Lambda_t=O_d(1)$). On $\tau=O_d(1)$ times $c\to-1/\Gamma_t$ while on $\tau=O_d(d)$ times $(m,q,b)\to(1/\Gamma_t,0,\tfrac12\log\tfrac{\omega}{1-\omega})$.
\end{enumerate}
\end{result}

\begin{figure}[h!]
    \centering
    \includegraphics[width=0.8\figwidth]{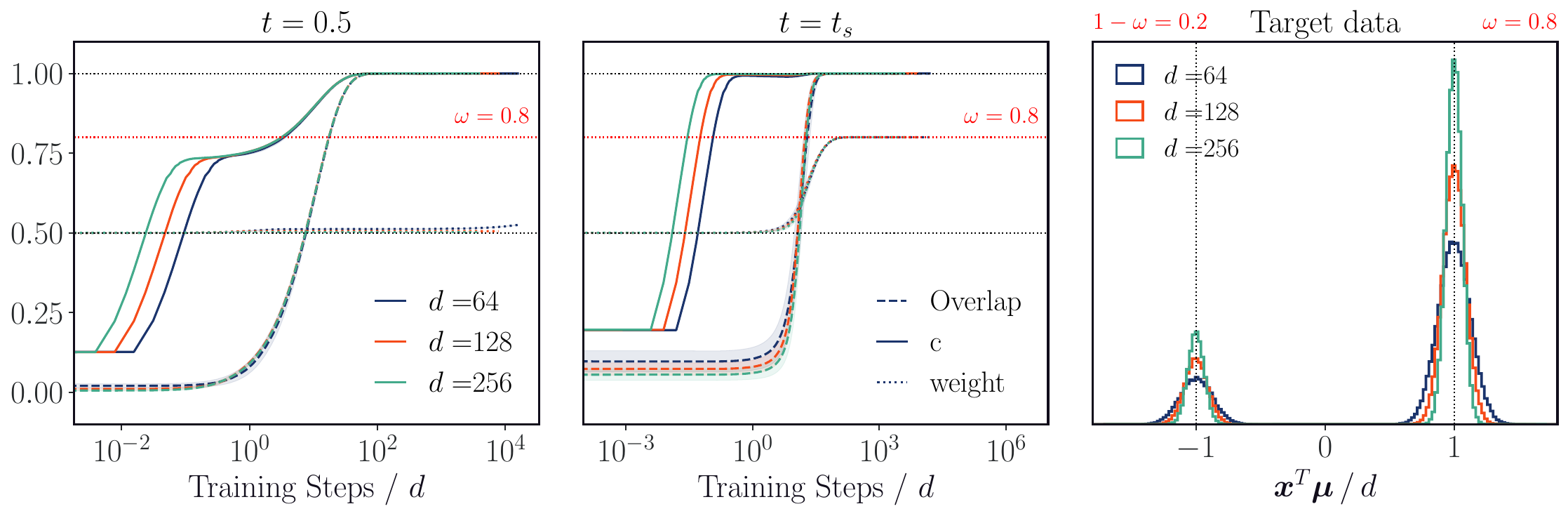}
    \caption{\textbf{Exact score parametrization for $(\mathcal{D}_1)$.} Overlap $m$ (dashed), skip connection $c$ (solid) and mode weight (dotted), against training steps rescaled by $d$, for the unbalanced GMM at $\omega=0.8$, different values of $d$ and two noising times, $t=0.5$ \textit{(left)} and $t=t_s$ \textit{(middle)}. 
    Typical bimodal histograms of projected samples are shown on the $(right)$ panel.
    In {\bf Regime~II} \textit{(left)} $c$ plateaus on $O(1)$ times and $m$ converges on $O(d)$ times, while the weight stays frozen at its initial value $1/2$.
    At {\bf speciation} \textit{(right)}, $c$ first reaches its final value and then weight and overlap rise together, on a shared $O(d)$ timescale.
    }
    \label{fig:exscore-unbalanced}
\end{figure}

On $\tau=O_d(1)$ times only the skip connection $c$ evolves; the model learns the score associated to $\mathcal{N}(0,(\Gamma_t+\alpha_t^2)\vI_d)$, the best unimodal approximation to the bimodal target.
Multimodality is then acquired on $\tau=O_d(d)$ times and in a regime-dependent fashion:
in Regime~II the network recovers the exact score \eqref{eq:unbalanced} up to its bias, so the direction is learned but the imbalance is not (Fig.~\ref{fig:exscore-unbalanced}, \textit{left}); at speciation, direction and imbalance are recovered jointly (Fig.~\ref{fig:exscore-unbalanced}, \textit{middle}).
 
\begin{result}[Training dynamics for $(\mathcal{D}_2)$ (\S\ref{app:hierarchical_exact})]
\label{thm:hierarchical}
Let $\kappa_1=\kappa$, $\kappa_2=1-\kappa$, $\vw_1(0)\in\mathbb S^{\kappa_1 d-1}$, $\vw_2(0)\in\mathbb S^{\kappa_2 d-1}$ and $c(0)=O_d(1)$. Let $(c,\vw_1,\vw_2)$ evolve under the GF \eqref{eq:grad_flow} for the score model \eqref{eq:score_model_4modes} at fixed $\Lambda_t$. In the $d\to\infty$ limit the $d$-dimensional flow closes on the summary statistics $(m_1,q_1,m_2,q_2,c)$, each pair $(m_i,q_i)$ obeying the system of Result~\ref{thm:unbalanced} at $\omega=1/2$, $b=0$ and rescaled SNR $\gamma_i^2=\kappa_i\Lambda_t$, the two systems being coupled only through $c$. Specifically:
\begin{enumerate}[label=(\alph*)]
    \item \textbf{Regime~II} ($\Lambda_t=O_d(d)$). On $\tau=O_d(1)$ times $c\to -1/(\Gamma_t+\alpha_t^2)$, while $(m_i,q_i)$ stay at initialization. On $\tau=O_d(d)$ times $c$ relaxes instantaneously to $c^*(m_1,q_1, m_2, q_2)$ while $(m_i,q_i)\to(1/\Gamma_t,0)$ on a $\kappa$-independent timescale.
    \item \textbf{Speciation} ($\Lambda_t=O_d(1)$). On $\tau=O_d(1)$ times $c\to-1/\Gamma_t$, the two systems decouple and, for $i=1,2$, $(m_i,q_i)\to(1/\Gamma_t,0)$ on $\kappa$-dependent timescales $\tau_i$. In the small $\kappa$ limit, $\tau_1/\tau_2\sim 1/\kappa$.
\end{enumerate}
\end{result}

\begin{figure}[h!]
    \centering
    \includegraphics[width=0.8\figwidth]{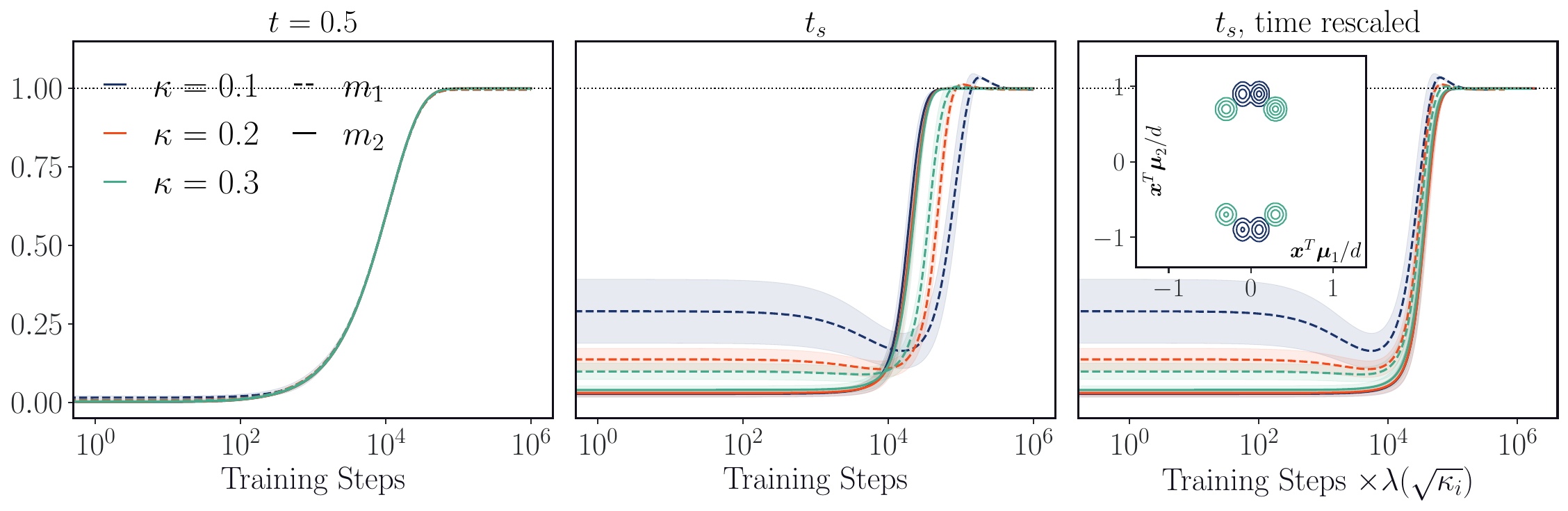}
    \caption{\textbf{Exact score parametrization for $(\mathcal{D}_2)$.} Block overlaps $m_1$ (dashed) and $m_2$ (solid), against training steps, for the quadrimodal GMM at $d=1024$, structure parameter $\kappa\in\{0.1,0.2,0.3\}$ and two noising times, $t=0.5$ \textit{(left)} and $t=t_s$ \textit{(middle and right)}. In {\bf Regime~II} \textit{(left)} the two overlaps are learned on the same training timescale, independently of $\kappa$. At {\bf speciation} \textit{(middle)} they evolve on different timescales with $\vmu_1$ being learned after $\vmu_2$. Rescaling each block's training time by its own closed-form rate $\lambda(\gamma_i)$ of \eqref{eq:mu_main}, with $\gamma_i^2=\kappa_i\Lambda_t\Gamma_t$ and $\kappa_1=\kappa,\ \kappa_2=1-\kappa$, collapses the curves \textit{(right)}. The inset shows $(\mathcal{D}_2)$ samples, projected on $(\vmu_1, \vmu_2)$. The smaller $\kappa$, the closer the modes pair up along $\vmu_1$ and the more hierarchical the target.
}
\label{fig:exscore-kappa}
\end{figure}

As in the unbalanced case, on $\tau=O_d(1)$ times the model only adapts its skip connection and learns the best unimodal approximation of the target. The two directions are then acquired on a $O_d(d)$ timescale, in a regime-dependent fashion. 
In Regime~II both overlaps evolve jointly, independently of the block asymmetry (Fig.~\ref {fig:exscore-kappa}, \textit{left}); at speciation however each block evolves at its own SNR set by $\kappa$. Consequently, learning is hierarchical, with the dominant direction $\vmu_2$ acquired before $\vmu_1$ (Fig.~\ref{fig:exscore-kappa}, \textit{middle}). 

Since the dynamics of $(m_i,q_i)$ are coupled, the rate of growth of $m_i$ drifts as $q_i$ relaxes, making its $\kappa$ dependence intractable. To proceed further, we impose the extra constraint of fixed $\vw_i$ norm.
\begin{resultbis}[Heuristic growth rate \S\ref{app:fixednorm}]
Constrain each $\vw_i$ to its target norm, $\lVert\vw_i\rVert=\alpha(t)\lVert\vmu_i\rVert$, i.e.\ $m_i^2+q_i=1$. The small $m_i$ linearized dynamics of Result~\ref{thm:hierarchical}(b) then reduces to the single ODE
\begin{align}
\label{eq:mu_main}
     \dot{m}_i = d^{-1}\lambda(\gamma_i)\,m_i+O(m_i^3),
\end{align}
with $\lambda(\gamma)$ given in closed form in \eqref{eq:Phi_expand_C} of \S\ref{app:fixednorm}.
\end{resultbis}

While this assumption is provably not exact, it is closed-form and recovers \ref{thm:hierarchical}(b) exactly as $\kappa\to0$. 
Consequently, we use $\tau_i\propto1/\lambda(\gamma_i)$ as the reference timescale against which training time is rescaled in Fig.~\ref{fig:exscore-kappa} and throughout. \looseness=-1

\subsection{Integrated loss}
\label{sec:integrated_loss}

Standard SI pipelines optimize the integrated loss \eqref{eq_DSM} rather than a single-time one, exposing the model to many noise levels at once. Since this loss is a weighted superposition of fixed-SNR denoising problems, Results~\ref{thm:unbalanced} and~\ref{thm:hierarchical} suggest that training depends on how much weight is placed on each SNR region, which we now make precise. \looseness=-1

Rewriting a denoising objective in terms of SNR rather than time is now standard \citep{kingma2023understanding}, and underlies state-of-the-art schedule designs \citep{karras2022elucidatingdesignspacediffusionbased, esser2024scaling}.
Switching from $t\mapsto\Lambda$ in the DSM loss \eqref{eq_DSM}, we collect every schedule $(\alpha, \beta)$ and weighting $w$ into a single effective SNR weight $w_{\mathrm{eff}}(\Lambda)=w(t(\Lambda))\lvert\dd t/\dd\Lambda\rvert$, which fully characterizes a given SI framework (\S\ref{app:integrated_snr}).
In light of the fixed-SNR analysis, most of the structural information is passed to the model around speciation: the relevant signal peaks at $\Lambda=O_d(1)$, and comparing pipelines amounts to comparing the weight they place around that scale.
Under this lens, standard Diffusion (DM) and Flow Matching (FM)\footnote{For Flow Matching the velocity regression must first be rewritten as a noise prediction, which contributes an extra Jacobian $1/\alpha^2(t)$ (\S\ref{app:integrated_snr}).} pipelines differ explicitly. For $\Lambda \ll d$,
\begin{align}
\label{eq:weff_main}
    w_{\mathrm{eff}}^{\mathrm{DM}}(\Lambda)\propto \Lambda^{-1}
    \qquad
    w_{\mathrm{eff}}^{\mathrm{FM}}(\Lambda)\propto \sqrt{d}\Lambda^{-3/2}.
\end{align}
Diffusion is scale-free and allocates the same denoising effort to every decade of SNR, exposing the network to a broad mixture of noise levels from Regime~II down past speciation.
Conversely, Flow Matching carries the extra factor $\sqrt{d/\Lambda}$ and tilts training towards low SNR.
Note that this concentration is not a design choice but a consequence of the canonical FM schedule itself.
We therefore expect the FM training dynamics to be dominated by speciation-specific effects, whereas Diffusion should display more mixed behavior.
In turn, this predicts distinct structure learning signatures, which we test on $(\mathcal{D}_1)$ and $(\mathcal{D}_2)$ by replacing the analytical score of \S\ref{sec:fixed_time} with a generic ResNet denoiser trained under both FM and DM effective SNR weighting functions (\S\ref{sec:dsm-sibling}).
Neither pipeline is expected to reproduce the fixed-SNR analysis exactly --- the network shares parameters across noise levels, so the fixed-SNR dynamics do not simply superpose --- but we still find qualitative agreement.

\begin{figure}[h!]
    \centering
    \begin{minipage}{1.0\figwidth}
        \begin{subfigure}[t]{0.495\linewidth}
            \centering
            \includegraphics[width=\linewidth]{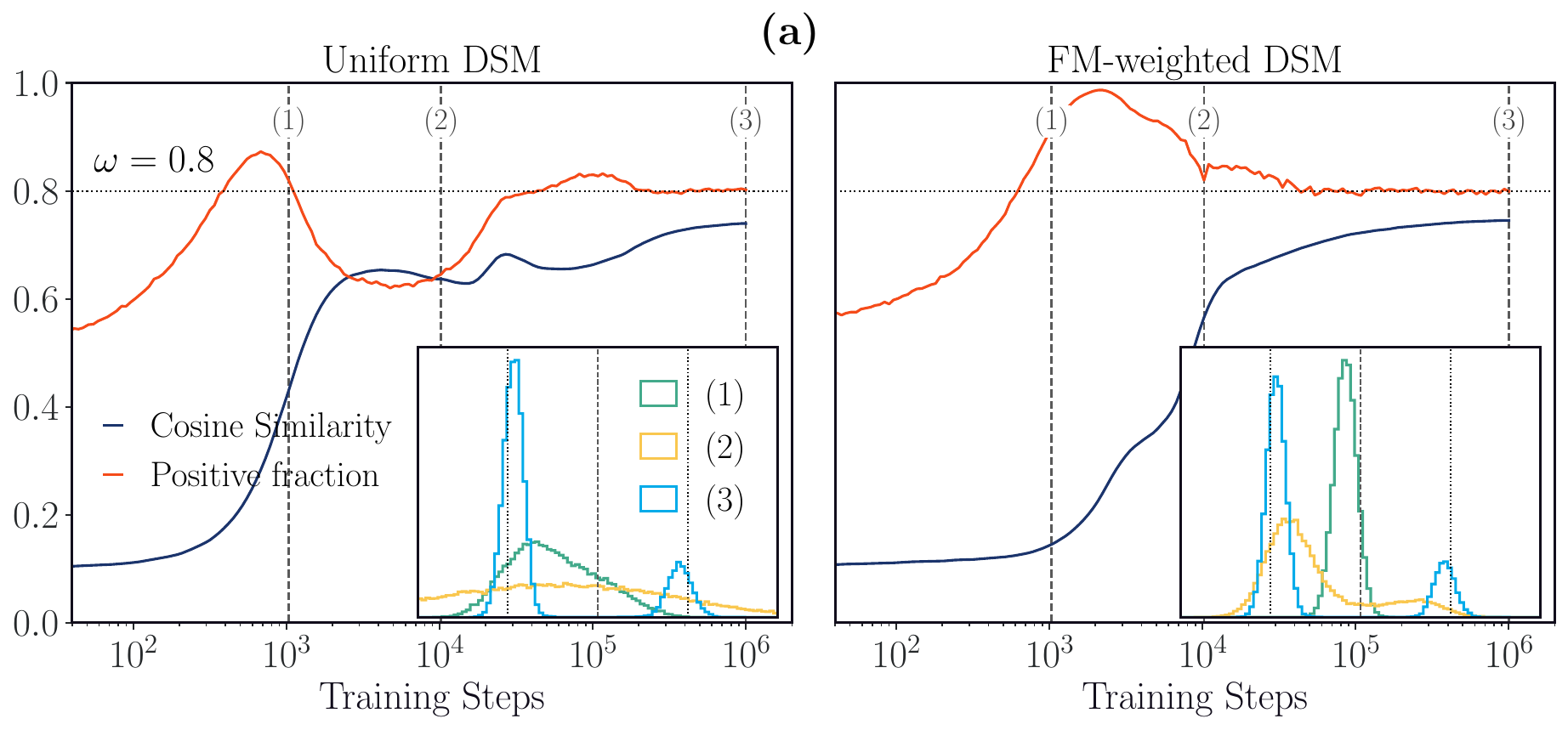}
            \phantomsubcaption
            \label{fig:gmm-resnet-weight-learning}
        \end{subfigure}
        \hfill
        \begin{subfigure}[t]{0.495\linewidth}
            \centering
            \includegraphics[width=\linewidth]{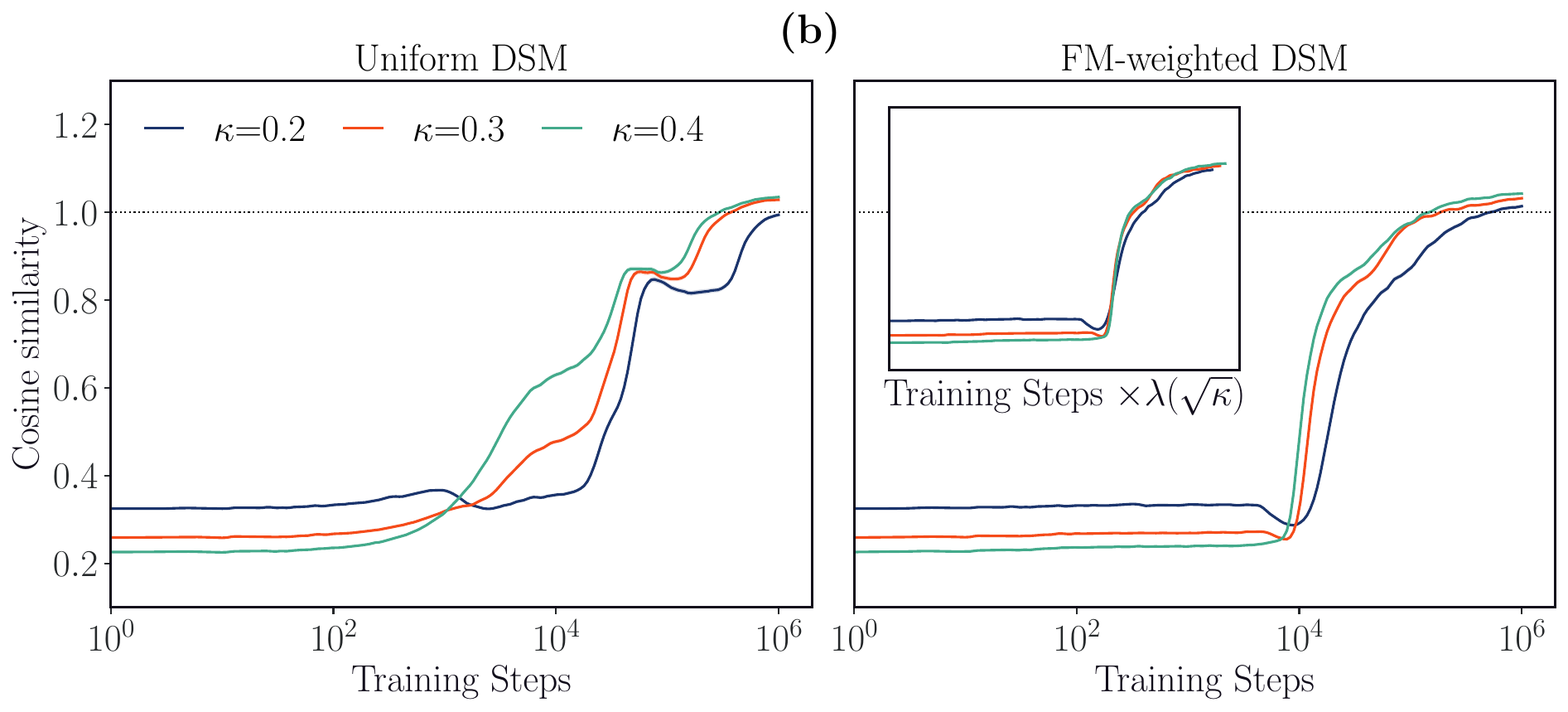}
            \phantomsubcaption
            \label{fig:gmm-resnet-struct-learning}
        \end{subfigure}
    \end{minipage} 
    \caption{
    \textbf{ResNet for GMMs} at $d=64$, under uniform DSM (\textit{left}) and FM-weighted DSM (\textit{right}) in each panel, against training steps.
    \textbf{(a)} $(\mathcal{D}_1)$ at $\omega=0.8$: cosine similarity with $\vmu$ (blue) and positive fraction of $p=\vx^T\vmu/\lVert\vmu\rVert^2$ (red); insets show the histogram of $p$ at the three marked steps.
    \textbf{(b)} $(\mathcal{D}_2)$: cosine similarity with $\vmu_1$ for $\kappa\in\{0.2,0.3,0.4\}$; the inset replots the FM curves against training steps rescaled by $\lambda(\sqrt{\kappa})$.
    }
    
    \label{fig:gmm-resnet}
\end{figure}

On $(\mathcal D_1)$ (Fig.~\ref{fig:gmm-resnet-weight-learning}), once two modes appear in the generated samples, uniform DSM first relaxes towards an almost even split, as predicted by Result~\ref{thm:unbalanced}(a), before converging to $\omega$; its earlier rise in positive fraction is spurious, reporting a displaced mean while the histograms are still unimodal at $(1)$. Under the FM weighting, weight and direction are instead acquired together, as at speciation. On $(\mathcal D_2)$ (Fig.~\ref{fig:gmm-resnet-struct-learning}), the overlap with the subdominant direction $\vmu_1$ separates with $\kappa$ under both weightings, more sharply under FM, for which rescaling training steps by $\lambda(\sqrt{\kappa})$ collapses the curves.

\section{Structure learning in complex data}
 
While synthetic GMMs allow for a clear interpretation of the structural features composing the data,
such identification in real world datasets is typically open to interpretation. Nevertheless we demonstrate that our theory
captures learning dynamics across data modalities and SI design choices through three numerical experiments of increasing complexity. 
Precise details on all numerical experiments are provided in appendix (\S\ref{app:numerical_details}).

\begin{figure}[h!]
    \centering
    \begin{minipage}{1.0\figwidth}
        \begin{subfigure}[t]{0.495\linewidth}
            \centering
            \includegraphics[width=\linewidth]{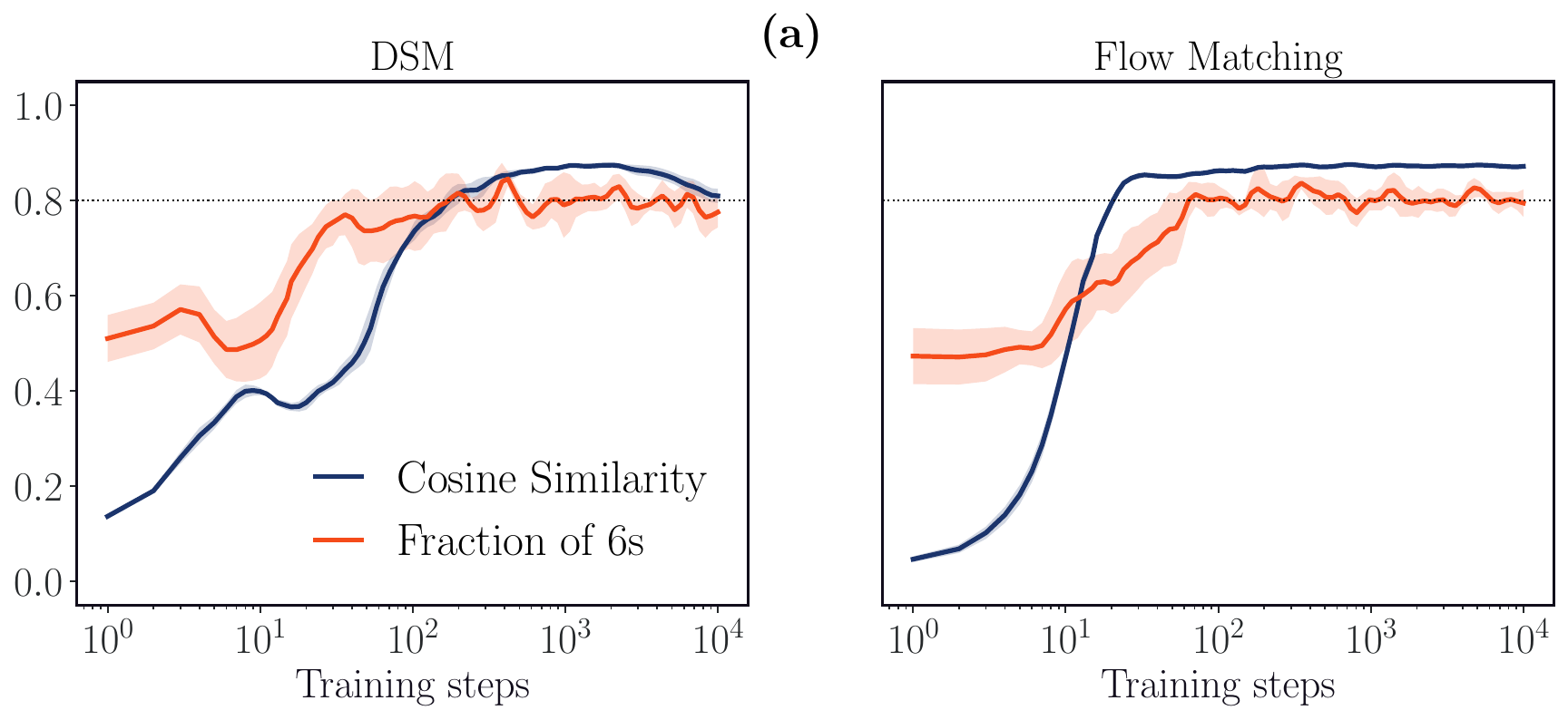}
            \phantomsubcaption
            \label{fig:mnist-weight-learning}
        \end{subfigure}
        \hfill
        \begin{subfigure}[t]{0.495\linewidth}
            \centering
            \includegraphics[width=\linewidth]{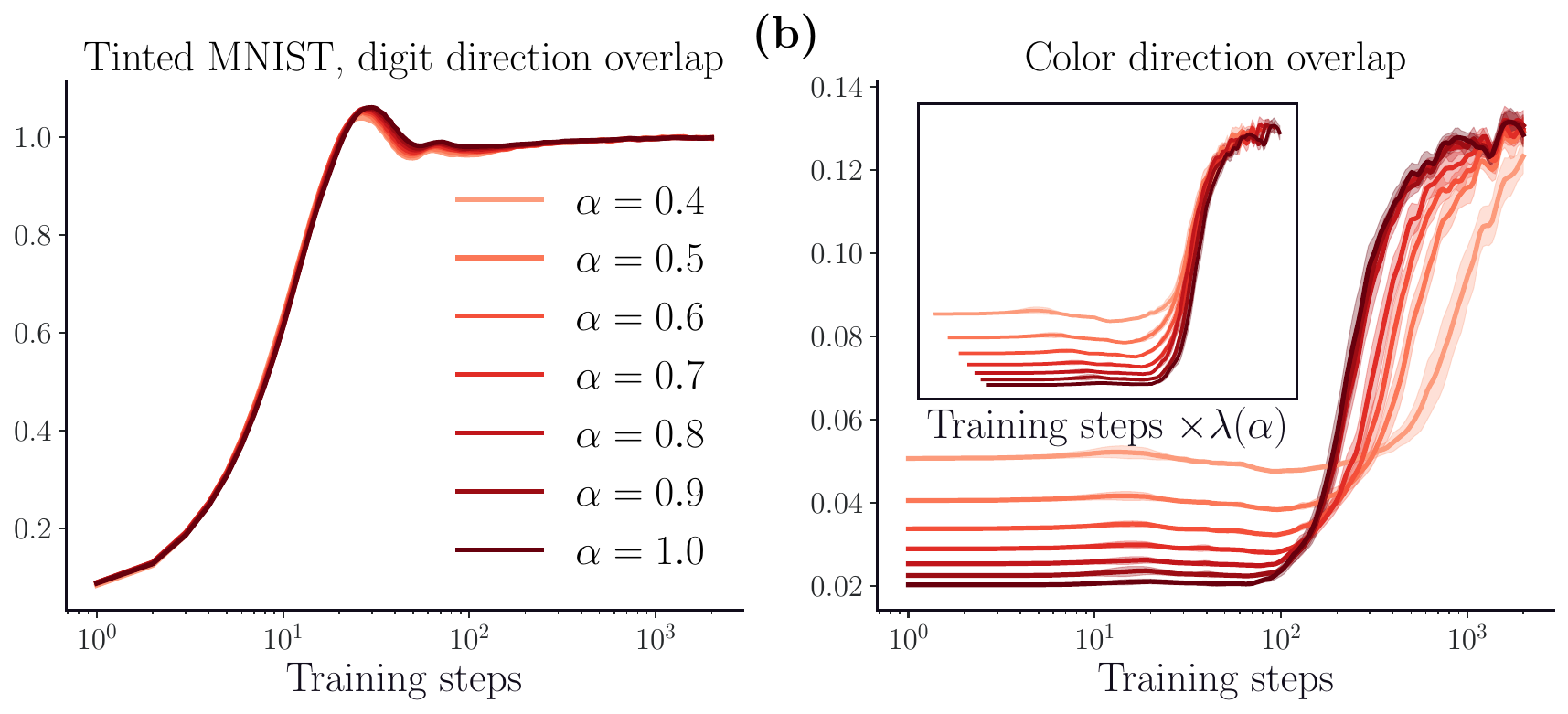}
            \phantomsubcaption
            \label{fig:mnist-hierarchical}
        \end {subfigure}
    \end{minipage}
    \caption{\textbf{Experiments on MNIST.}
    \textbf{(a)} Cosine similarity with the digit direction (blue) and fraction of 6s (orange), against training steps, for MNIST restricted to 3s and 6s, of which a fraction $\omega=0.8$ are 6s (dotted line), under DSM \textit{(left)} and FM \textit{(right)}. Both objectives acquire direction and proportion within a single decade of training, MNIST offering too short an SNR range for the two weightings to separate.
    \textbf{(b)} Overlap with the digit direction \textit{(left)} and with the color direction \textit{(right)}, against training steps, for tinted MNIST trained with FM at tint intensities $\alpha$, the dataset being built so that the squared amplitude of the color mode scales as $\alpha^2$. The digit direction is learned at a rate nearly insensitive to $\alpha$, while the color direction separates with it; rescaling training time by $\lambda(\alpha)$ collapses the color curves (inset). \looseness=-1
    }
    \label{fig:mnist}
\end{figure}

\paragraph*{Weight learning in unbalanced MNIST.} The paradigmatic multi-class MNIST dataset offers a natural route to test coordinated learning of class index and class proportion. To mimic a bimodal distribution we prune MNIST to two classes $3/6$ with relative tunable proportion $\omega$. We compare on Fig. \ref{fig:mnist-weight-learning} out-of-the-box DSM and FM implementations for a fixed imbalance $\omega=0.8$ in favor of the 6s and use a pretrained digit classifier for evaluating empirical class proportion.

Under Flow Matching, class proportion and direction evolve jointly: as denoising progresses, the fraction of blurry images classified as 6s ramps up to $\omega$ concurrently with increasing image crispness, as expected of a pipeline overweighting low SNR regions.
Unlike the controlled ResNet experiments discussed above, where only $w$ changes, however, the comparison with DSM is inconclusive: proportion and direction appear to be learned on the same timescale under both objectives. We attribute this to the weaker mode separation of unbalanced MNIST, which shortens the relevant SNR range distinguishing the two weighting schemes (the two out-of-the-box pipelines also differ in more than $w$, \S\ref{sec:training-mnist}).

\paragraph*{Learning hierarchy in (Tinted MNIST/FM).} 
To investigate hierarchical feature emergence during training we design a new dataset by adding a digit-orthogonal mode to the $3/6$ pruned MNIST. Specifically, we randomly colorize each image in either red or green with tunable intensity $\alpha \in (0,1]$; see Fig. \ref{fig:app-tint-ex} for representative samples and appendix \ref{sec:data-mnist} for details on the dataset construction.
Importantly, the resulting dataset is reminiscent of $\mathcal{D}_2$, with orthogonal digit and color directions.
To match the GMM analysis, the squared amplitude of the color direction scales as $\alpha^2$ and can thus be made arbitrarily small. \looseness=-1
 
Within the FM pipeline, training is dominated by the speciation region; consequently, digit learning occurs on a single $\alpha$-independent timescale, whereas the color direction is acquired on an $\alpha$ dependent one, its onset spanning close to a decade across the sweep (Fig.~\ref{fig:mnist-hierarchical}).
We emphasize that even on this complex, image-based dataset, the GMM rescaling ansatz \eqref{eq:mu_main} still quantitatively captures the onset of color mode emergence, provided the experiment is matched to the single SNR analysis by identifying the small mode amplitude $\lvert\vmu_1\rvert \propto \alpha$.
 
\begin{figure}[h!]
   \centering
   \includegraphics[width=0.9\figwidth]{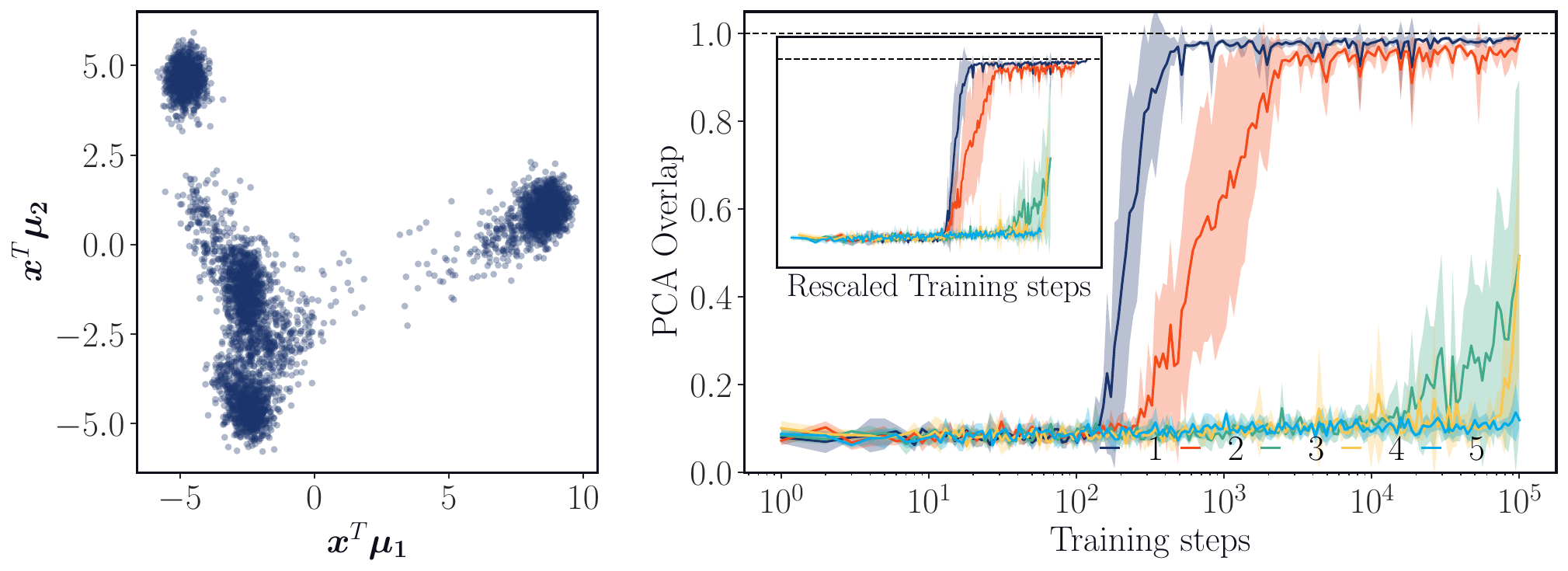}
   \caption{
   \textbf{Sequential feature learning on HGD.} \textit{(Left)} HGD samples projected onto the first two PCA eigenvectors, showing a complex multimodal structure. \textit{(Right)} Overlap of the generated samples with the leading PCA components $\vmu_i$, $i=1,\dots,5$, against training steps, under Masked Discrete Diffusion: the directions are recovered one after the other, and rescaling training time by the closed-form rate $\lambda(\sqrt{v_i})$ of \eqref{eq:mu_main}, with $v_i$ the variance explained along the $i$-th reference PCA direction, collapses the two leading ones (inset).  \looseness=-1
   }
    \label{fig:mdlm-humgen}
\end{figure}

\paragraph*{Sequentially learning structured features in genomic data (HGD/MDLM).} 
Departing from synthetic datasets, we finally consider haplotype data (binary sequences) from the Human Genome Dataset (HGD), following \citet{Bachtis_2024}.
To test the scope of our theory, we consider yet another SI framework, Masked Discrete Diffusion, adapting the original implementation of \citet{sahoo2024mdlm} to the binary data at hand (see App. \ref{app:mdlm-humgen} for implementation details).
Unlike in the GMM and tinted MNIST cases, feature structure in the HGD is not made obvious by construction. To assess sequential learning of data features we pre-compute the training data's PCA eigenvectors and track the statistics of inferred samples projected on these fixed directions, shown in Fig. \ref{fig:mdlm-humgen}. \looseness=-1

The complexity of the HGD is made apparent in the left panel of Fig. \ref{fig:mdlm-humgen}, where the first and second eigenvector projections display clear multi-modality.
The right panel shows that the generated data acquires the top PCA eigendirections sequentially, the leading two being separated by close to half a decade of training while the subsequent ones only begin to rise at the end of the run.
Following the single SNR analysis, we rescale training time by $\lambda(\sqrt{v_k})$, with $v_k$ the variance explained along direction $k$ (App. \ref{app:mdlm-humgen}). The collapse holds for the two leading directions, precisely those along which the projected data is visibly multimodal (left panel), and degrades beyond: the smaller eigendirections carry no resolved modes, so the mode amplitude the ansatz rests on has no counterpart there.
The MDLM loss is moreover scale free (App. \ref{app:mdlm_snr}), which, as in the ResNet experiment of Fig. \ref{fig:gmm-resnet-struct-learning}, does not prevent sequential emergence. \looseness=-1

\section{Conclusion and future work}

Reading a training objective through its effective SNR weighting turns an
arbitrary design choice into a predictive one: which features a model resolves,
and when, follows from where that weighting places its mass relative to the
speciation scale. Gaussian mixtures make this concrete. Training deep in
Regime~II of \citet{biroli_2024} recovers every mode direction on a common
timescale but leaves the relative weights untouched, however long one trains.
Training around speciation instead acquires weights and directions jointly, and
separates the directions themselves at rates set by their relative amplitudes.
The same phenomenology holds for networks trained on unbalanced MNIST and, quantitatively, on tinted MNIST and genomic haplotype data.
\looseness=-1

The same statement, read backwards, is a design principle. Rather than inheriting a weighting schedule, one could tailor it to the structure of a given target, so as to acquire a chosen feature first or to reach a desired structure within a fixed compute budget. We leave this to future work. \looseness=-1

\section*{Acknowledgment.}
GB acknowledges support from the French government under the management of ANR: PEPR-IA (project MAGICALL ANR-25-PEIA-0004) and PR[AI]RIE-PSAI (ANR-23-IACL-
0008). JK acknowledges financial support from the postdoctoral Junior Research and Teaching Laplace chair, funded by Capital Fund Management and the \'Ecole Normale Supérieure.

\bibliography{biblio_arxiv}

@inproceedings{sohl-dickstein_15,
  author    = {Sohl-Dickstein, Jascha and Weiss, Eric and Maheswaranathan, Niru and Ganguli, Surya},
  title     = {Deep Unsupervised Learning using Nonequilibrium Thermodynamics},
  booktitle = {Proceedings of the 32nd International Conference on Machine Learning},
  volume    = {37},
  pages     = {2256--2265},
  year      = {2015},
  url       = {https://proceedings.mlr.press/v37/sohl-dickstein15.html},
}

@inproceedings{Ho_DDPM,
  author    = {Ho, Jonathan and Jain, Ajay and Abbeel, Pieter},
  title     = {Denoising Diffusion Probabilistic Models},
  booktitle = {Advances in Neural Information Processing Systems},
  volume    = {33},
  pages     = {6840--6851},
  year      = {2020},
  url       = {https://proceedings.neurips.cc/paper_files/paper/2020/file/4c5bcfec8584af0d967f1ab10179ca4b-Paper.pdf},
}

@inproceedings{song2021scorebased,
  author    = {Song, Yang and Sohl-Dickstein, Jascha and Kingma, Diederik P. and Kumar, Abhishek and Ermon, Stefano and Poole, Ben},
  title     = {Score-Based Generative Modeling through Stochastic Differential Equations},
  booktitle = {International Conference on Learning Representations},
  year      = {2021},
  url       = {https://arxiv.org/abs/2011.13456},
}

@inproceedings{lipman2023flow,
  author    = {Lipman, Yaron and Chen, Ricky T. Q. and Ben-Hamu, Heli and Nickel, Maximilian and Le, Matthew},
  title     = {Flow Matching for Generative Modeling},
  booktitle = {International Conference on Learning Representations},
  year      = {2023},
  url       = {https://arxiv.org/abs/2210.02747},
}

@inproceedings{albergo2023stochastic,
  author    = {Albergo, Michael S. and Vanden-Eijnden, Eric},
  title     = {Building Normalizing Flows with Stochastic Interpolants},
  booktitle = {International Conference on Learning Representations},
  year      = {2023},
  url       = {https://arxiv.org/abs/2209.15571},
}

@article{Albergo2025unifying,
  author  = {Albergo, Michael S. and Boffi, Nicholas M. and Vanden-Eijnden, Eric},
  title   = {Stochastic Interpolants: A Unifying Framework for Flows and Diffusions},
  journal = {Journal of Machine Learning Research},
  volume  = {26},
  number  = {209},
  pages   = {1--80},
  year    = {2025},
  url     = {https://jmlr.org/papers/v26/23-1605.html},
}

@article{hyvarinen_05,
  author  = {Hyv{\"a}rinen, Aapo},
  title   = {Estimation of Non-Normalized Statistical Models by Score Matching},
  journal = {Journal of Machine Learning Research},
  volume  = {6},
  number  = {24},
  pages   = {695--709},
  year    = {2005},
  url     = {https://jmlr.org/papers/v6/hyvarinen05a.html},
}

@article{Vincent_2011,
  author  = {Vincent, Pascal},
  title   = {A Connection Between Score Matching and Denoising Autoencoders},
  journal = {Neural Computation},
  volume  = {23},
  number  = {7},
  pages   = {1661--1674},
  year    = {2011},
  url     = {https://doi.org/10.1162/NECO_a_00142},
}

@article{robbins1951stochastic,
  author  = {Robbins, Herbert and Monro, Sutton},
  title   = {A Stochastic Approximation Method},
  journal = {The Annals of Mathematical Statistics},
  volume  = {22},
  number  = {3},
  pages   = {400--407},
  year    = {1951},
  url     = {https://doi.org/10.1214/aoms/1177729586},
}

@article{Anderson_1982,
  author  = {Anderson, Brian D. O.},
  title   = {Reverse-time diffusion equation models},
  journal = {Stochastic Processes and their Applications},
  volume  = {12},
  number  = {3},
  pages   = {313--326},
  year    = {1982},
  url     = {https://doi.org/10.1016/0304-4149(82)90051-5},
}

@article{haussmann_1986,
  author  = {Haussmann, U. G. and Pardoux, E.},
  title   = {Time Reversal of Diffusions},
  journal = {The Annals of Probability},
  volume  = {14},
  number  = {4},
  pages   = {1188--1205},
  year    = {1986},
  url     = {https://doi.org/10.1214/aop/1176992362},
}

@inproceedings{choi2022perception,
  author    = {Choi, Jooyoung and Lee, Jungbeom and Shin, Chaehun and Kim, Sungwon and Kim, Hyunwoo and Yoon, Sungroh},
  title     = {Perception Prioritized Training of Diffusion Models},
  booktitle = {IEEE/CVF Conference on Computer Vision and Pattern Recognition (CVPR)},
  pages     = {11462--11471},
  year      = {2022},
  url       = {https://arxiv.org/abs/2204.00227},
}

@inproceedings{hang2023efficient,
  author    = {Hang, Tiankai and Gu, Shuyang and Li, Chen and Bao, Jianmin and Chen, Dong and Hu, Han and Geng, Xin and Guo, Baining},
  title     = {Efficient Diffusion Training via Min-{SNR} Weighting Strategy},
  booktitle = {IEEE/CVF International Conference on Computer Vision (ICCV)},
  year      = {2023},
  url       = {https://arxiv.org/abs/2303.09556},
}

@inproceedings{kingma2023understanding,
  author    = {Kingma, Diederik P. and Gao, Ruiqi},
  title     = {Understanding Diffusion Objectives as the {ELBO} with Simple Data Augmentation},
  booktitle = {Advances in Neural Information Processing Systems},
  volume    = {36},
  year      = {2023},
  url       = {https://arxiv.org/abs/2303.00848},
}

@inproceedings{esser2024scaling,
  author    = {Esser, Patrick and Kulal, Sumith and Blattmann, Andreas and Entezari, Rahim and M{\"u}ller, Jonas and Saini, Harry and Levi, Yam and Lorenz, Dominik and Sauer, Axel and Boesel, Frederic and Podell, Dustin and Dockhorn, Tim and English, Zion and Lacey, Kyle and Goodwin, Alex and Marek, Yannik and Rombach, Robin},
  title     = {Scaling Rectified Flow Transformers for High-Resolution Image Synthesis},
  booktitle = {Proceedings of the 41st International Conference on Machine Learning},
  year      = {2024},
  url       = {https://arxiv.org/abs/2403.03206},
}

@misc{karras2022elucidatingdesignspacediffusionbased,
  author = {Karras, Tero and Aittala, Miika and Aila, Timo and Laine, Samuli},
  title  = {Elucidating the Design Space of Diffusion-Based Generative Models},
  year   = {2022},
  url    = {https://arxiv.org/abs/2206.00364},
}

@misc{aranguri2025optimizingnoiseschedulesgenerative,
  author = {Aranguri, Santiago and Biroli, Giulio and M{\'e}zard, Marc and Vanden-Eijnden, Eric},
  title  = {Optimizing Noise Schedules of Generative Models in High Dimensions},
  year   = {2025},
  url    = {https://arxiv.org/abs/2501.00988},
}

@misc{aranguri2025phaseaware,
  author = {Aranguri, Santiago and Insulla, Francesco},
  title  = {Phase-aware Training Schedule Simplifies Learning in Flow-Based Generative Models},
  year   = {2024},
  url    = {https://arxiv.org/abs/2412.07972},
}

@inproceedings{Raya_2023,
  author    = {Raya, Gabriel and Ambrogioni, Luca},
  title     = {Spontaneous Symmetry Breaking in Generative Diffusion Models},
  booktitle = {Advances in Neural Information Processing Systems},
  volume    = {36},
  pages     = {66377--66389},
  year      = {2023},
  url       = {https://proceedings.neurips.cc/paper_files/paper/2023/file/d0da30e312b75a3fffd9e9191f8bc1b0-Paper-Conference.pdf},
}

@article{Biroli_2023,
  author  = {Biroli, Giulio and M{\'e}zard, Marc},
  title   = {Generative diffusion in very large dimensions},
  journal = {Journal of Statistical Mechanics: Theory and Experiment},
  volume  = {2023},
  number  = {9},
  pages   = {093402},
  year    = {2023},
  url     = {https://doi.org/10.1088/1742-5468/acf8ba},
}

@article{biroli_2024,
  author  = {Biroli, Giulio and Bonnaire, Tony and de Bortoli, Valentin and M{\'e}zard, Marc},
  title   = {Dynamical regimes of diffusion models},
  journal = {Nature Communications},
  volume  = {15},
  number  = {1},
  pages   = {9957},
  year    = {2024},
  url     = {https://doi.org/10.1038/s41467-024-54281-3},
}

@article{sclocchi2025phase,
  author  = {Sclocchi, Antonio and Favero, Alessandro and Wyart, Matthieu},
  title   = {A Phase Transition in Diffusion Models Reveals the Hierarchical Nature of Data},
  journal = {Proceedings of the National Academy of Sciences},
  volume  = {122},
  number  = {1},
  pages   = {e2408799121},
  year    = {2025},
  url     = {https://doi.org/10.1073/pnas.2408799121},
}

@inproceedings{Bachtis_2024,
  author    = {Bachtis, Dimitrios and Biroli, Giulio and Decelle, Aur{\'e}lien and Seoane, Beatriz},
  title     = {Cascade of Phase Transitions in the Training of Energy-Based Models},
  booktitle = {Advances in Neural Information Processing Systems},
  volume    = {37},
  pages     = {55591--55619},
  year      = {2024},
  url       = {https://proceedings.neurips.cc/paper_files/paper/2024/file/648a5a590ca6f2bb5de53f938e230160-Paper-Conference.pdf},
}

@misc{li2024criticalwindowsnonasymptotictheory,
  author = {Li, Marvin and Chen, Sitan},
  title  = {Critical Windows: Non-Asymptotic Theory for Feature Emergence in Diffusion Models},
  year   = {2024},
  url    = {https://arxiv.org/abs/2403.01633},
}

@misc{ambrogioni2023statistical,
  author = {Ambrogioni, Luca},
  title  = {The Statistical Thermodynamics of Generative Diffusion Models},
  year   = {2023},
  url    = {https://arxiv.org/abs/2310.17467},
}

@misc{gagneux2025generationphases,
  author = {Gagneux, Anne and Martin, S{\'e}gol{\`e}ne and Gribonval, R{\'e}mi and Massias, Mathurin},
  title  = {The Generation Phases of Flow Matching: A Denoising Perspective},
  year   = {2025},
  url    = {https://arxiv.org/abs/2510.24830},
}

@inproceedings{cui2024,
  author    = {Cui, Hugo and Krzakala, Florent and Vanden-Eijnden, Eric and Zdeborov{\'a}, Lenka},
  title     = {Analysis of Learning a Flow-Based Generative Model from Limited Sample Complexity},
  booktitle = {International Conference on Learning Representations},
  pages     = {51929--51955},
  year      = {2024},
  url       = {https://proceedings.iclr.cc/paper_files/paper/2024/file/e45a448dfa778f6d62729a7bc8633c06-Paper-Conference.pdf},
}

@inproceedings{cui2025,
  author    = {Cui, Hugo and Pehlevan, Cengiz and Lu, Yue},
  title     = {A Solvable Model of Learning Generative Diffusion: Theory and Insights},
  booktitle = {Advances in Neural Information Processing Systems},
  volume    = {38},
  pages     = {5253--5296},
  year      = {2025},
  url       = {https://proceedings.neurips.cc/paper_files/paper/2025/file/082d3d795520c43214da5123e56a3a34-Paper-Conference.pdf},
}

@inproceedings{wang2026an,
  author    = {Wang, Binxu and Pehlevan, Cengiz},
  title     = {An Analytical Theory of Spectral Bias in the Learning Dynamics of Diffusion Models},
  booktitle = {Advances in Neural Information Processing Systems},
  year      = {2026},
  url       = {https://openreview.net/forum?id=SDhOClkyqC},
}

@misc{george2026denoisingscorematchingrandom,
  author = {George, Anand Jerry and Veiga, Rodrigo and Macris, Nicolas},
  title  = {Denoising Score Matching with Random Features: Insights on Diffusion Models from Precise Learning Curves},
  year   = {2026},
  url    = {https://arxiv.org/abs/2502.00336},
}

@misc{bardone2026theorylearningdatastatistics,
  author = {Bardone, Lorenzo and Merger, Claudia and Goldt, Sebastian},
  title  = {A Theory of Learning Data Statistics in Diffusion Models, from Easy to Hard},
  year   = {2026},
  url    = {https://arxiv.org/abs/2603.12901},
}

@misc{nicoletti2026interplaydatastructureimbalance,
  author = {Nicoletti, Flavio and Ma, Chenxiao and Ventura, Enrico and Saglietti, Luca and Sarao Mannelli, Stefano},
  title  = {The Interplay of Data Structure and Imbalance in the Learning Dynamics of Diffusion Models},
  year   = {2026},
  url    = {https://arxiv.org/abs/2605.06367},
}

@inproceedings{dhariwal2021diffusionmodelsbeatgans,
  author    = {Dhariwal, Prafulla and Nichol, Alexander Quinn},
  title     = {Diffusion Models Beat {GAN}s on Image Synthesis},
  booktitle = {Advances in Neural Information Processing Systems},
  year      = {2021},
  url       = {https://openreview.net/forum?id=OU98jZWS3x_},
}

@inproceedings{Rombach_2022,
  author    = {Rombach, Robin and Blattmann, Andreas and Lorenz, Dominik and Esser, Patrick and Ommer, Bj{\"o}rn},
  title     = {High-Resolution Image Synthesis with Latent Diffusion Models},
  booktitle = {IEEE/CVF Conference on Computer Vision and Pattern Recognition (CVPR)},
  pages     = {10674--10685},
  year      = {2022},
  url       = {https://doi.org/10.1109/CVPR52688.2022.01042},
}

@inproceedings{yang2025cogvideox,
  author    = {Yang, Zhuoyi and Teng, Jiayan and Zheng, Wendi and Ding, Ming and Huang, Shiyu and Xu, Jiazheng and Yang, Yuanming and Hong, Wenyi and Zhang, Xiaohan and Feng, Guanyu and Yin, Da and Zhang, Yuxuan and Wang, Weihan and Cheng, Yean and Xu, Bin and Gu, Xiaotao and Dong, Yuxiao and Tang, Jie},
  title     = {{CogVideoX}: Text-to-Video Diffusion Models with an Expert Transformer},
  booktitle = {International Conference on Learning Representations},
  year      = {2025},
  url       = {https://openreview.net/forum?id=LQzN6TRFg9},
}

@misc{sora2024,
  author = {Liu, Yixin and Zhang, Kai and Li, Yuan and Yan, Zhiling and Gao, Chujie and Chen, Ruoxi and Yuan, Zhengqing and Huang, Yue and Sun, Hanchi and Gao, Jianfeng and He, Lifang and Sun, Lichao},
  title  = {{Sora}: A Review on Background, Technology, Limitations, and Opportunities of Large Vision Models},
  year   = {2024},
  url    = {https://arxiv.org/abs/2402.17177},
}

@inproceedings{sahoo2024mdlm,
  author    = {Sahoo, Subham Sekhar and Arriola, Marianne and Schiff, Yair and Gokaslan, Aaron and Marroquin, Edgar and Chiu, Justin T. and Rush, Alexander and Kuleshov, Volodymyr},
  title     = {Simple and Effective Masked Diffusion Language Models},
  booktitle = {Advances in Neural Information Processing Systems},
  year      = {2024},
  url       = {https://arxiv.org/abs/2406.07524},
}

@article{noe2019boltzmann,
  author  = {No{\'e}, Frank and Olsson, Simon and K{\"o}hler, Jonas and Wu, Hao},
  title   = {Boltzmann Generators: Sampling Equilibrium States of Many-Body Systems with Deep Learning},
  journal = {Science},
  volume  = {365},
  number  = {6457},
  pages   = {eaaw1147},
  year    = {2019},
  url     = {https://doi.org/10.1126/science.aaw1147},
}

@article{gabrie2022adaptive,
  author  = {Gabri{\'e}, Marylou and Rotskoff, Grant M. and Vanden-Eijnden, Eric},
  title   = {Adaptive {M}onte {C}arlo Augmented with Normalizing Flows},
  journal = {Proceedings of the National Academy of Sciences},
  volume  = {119},
  number  = {10},
  pages   = {e2109420119},
  year    = {2022},
  url     = {https://doi.org/10.1073/pnas.2109420119},
}

@misc{grenioux2025amorphous,
  author = {Grenioux, Louis and Galliano, Leonardo and Berthier, Ludovic and Biroli, Giulio and Gabri{\'e}, Marylou},
  title  = {Boltzmann Generators for Amorphous Particle Systems},
  year   = {2025},
  url    = {https://arxiv.org/abs/2512.16607},
}

@misc{rehman2026autoregressive,
  author = {Rehman, Danyal and Tan, Charlie B. and Bengio, Yoshua and Bose, Avishek Joey and Tong, Alexander},
  title  = {Autoregressive {B}oltzmann Generators},
  year   = {2026},
  note   = {ICML 2026 (spotlight)},
  url    = {https://arxiv.org/abs/2606.27361},
}

@article{Behjoo_2025,
   title={U-Turn Diffusion},
   volume={27},
   ISSN={1099-4300},
   url={http://dx.doi.org/10.3390/e27040343},
   DOI={10.3390/e27040343},
   number={4},
   journal={Entropy},
   publisher={MDPI AG},
   author={Behjoo, Hamidreza and Chertkov, Michael},
   year={2025},
   month=Mar, pages={343} }

\newpage

\begin{center}
    {\LARGE Weighting Schedules Govern What and When Score-Based Generative Models Learn from Multimodal Data \\ \vspace{1ex} {\Large \bf Supplementary Material (SM)}
    \\ \vspace{3ex} }{\large J\'er\'emie Klinger, Rapha\"el Urfin, Giulio Biroli, Marylou Gabri\'e}
\end{center}

\appendix

\section{Details on the theoretical results}

\subsection{Dynamical regimes and signal-to-noise ratio}
\label{app:speciation}
In this section we adapt the computation of \citet{biroli_2024, aranguri2025optimizingnoiseschedulesgenerative} to stochastic interpolants, in order to identify the speciation time i.e. the time at which the generative trajectory commits to one of the modes of the target. Let us consider as a target distribution a bimodal symmetric isotropic Gaussian distribution      
\begin{align}
    P_0=\frac{1}{2}\mathcal{N}(\vmu, \sigma^2\vI_d)+\frac{1}{2}\mathcal{N}(-\vmu, \sigma^2\vI_d),\qquad \lVert \vmu\rVert^2=d.
\end{align}
Since the stochastic interpolant at time $t$ is $\vx_t=\alpha(t)\vx_0+\beta(t)\vxi$, the marginal $P_t$ is also a bimodal symmetric isotropic mixture,
\begin{align}
    P_t=\tfrac12\mathcal{N}(\alpha(t)\vmu, \Gamma_t\vI_d)+\tfrac12\mathcal{N}(-\alpha(t)\vmu, \Gamma_t\vI_d),\qquad \Gamma_t=\alpha^2(t)\sigma^2+\beta^2(t),
\end{align}
whose score reads
\begin{align}
\label{eq:score_sym}
    \nabla_{\vx}\log P_t(\vx)=-\frac{\vx}{\Gamma_t}+\frac{\alpha(t)\vmu}{\Gamma_t}\tanh\left(\frac{\alpha(t)\,\vmu^T\vx}{\Gamma_t}\right).
\end{align}

\paragraph*{Signal-to-noise ratio.} 
We study the evolution, during the generative dynamics, of the overlap
\begin{align}
\label{eq:q_def}
    q_t=\frac{\vmu^T\vx_t}{\sqrt{d}}.
\end{align}
Writing $\vx_0=s\vmu+\sigma\vz$, with $s=\pm1$ the label of the mode from which the sample is drawn and $\vz\sim\mathcal{N}(0,\vI_d)$, and projecting $\vx_t=\alpha(t)\vx_0+\beta(t)\vxi$ on $\vmu/\sqrt{d}$ gives
\begin{align}
\label{eq:q_decomp}
    q_t=\underbrace{s\,\alpha(t)\sqrt{d}}_{\text{signal}}\;+\;\underbrace{\alpha(t)\,\sigma\,g_1+\beta(t)\,g_2}_{\text{noise}},
    \qquad g_1,g_2\overset{\text{iid}}{\sim}\mathcal{N}(0,1),
\end{align}
where we used $\lVert\vmu\rVert^2=d$ and the fact that $\vmu^T\vz/\sqrt{d}$ and $\vmu^T\vxi/\sqrt{d}$ are standard Gaussians. The two noise contributions are independent, of total variance $\alpha^2(t)\sigma^2+\beta^2(t)=\Gamma_t$.
In turn, we define the time-dependent signal-to-noise ratio $\Lambda_t$ as
\begin{align}
\label{eq:snr_def}
    \Lambda_t=\frac{\big(\mathbb{E}\left[q_t\mid s\right]\big)^2}{\mathrm{Var}\left(q_t\mid s\right)}
    =\frac{\alpha^2(t)\lVert\vmu\rVert^2}{\Gamma_t}
    =\frac{\alpha^2(t)\,d}{\alpha^2(t)\sigma^2+\beta^2(t)},
\end{align}
where $\Lambda_t$ decreases from $\Lambda_t=d/\sigma^2$ at $t=0$ to $\Lambda_t=0$ at $t=\tmax$.

\paragraph*{Effective potential.} Project the generative ODE \eqref{eq:gen_ode}
\begin{align}
    \dot \vx(t)=\vb(\vx(t),t),\qquad \vb(\vx,t)=\frac{\dot \alpha(t)}{\alpha(t)}\vx-\left(\dot \beta(t)-\frac{\dot \alpha(t)}{\alpha(t)}\beta(t)\right)\beta(t)\nabla_{\vx}\log P_t(\vx)
\end{align}
onto $\vmu/\sqrt{d}$. Using \eqref{eq:score_sym} and $\vmu^T\vx=\sqrt{d}\,q$ yields a closed equation for $q$,
\begin{align}
\label{eq:q_potential_flow}
-\frac{\dd q_t}{\dd t}=-\frac{\partial V_{\mathrm{eff}}}{\partial q}(q_t),
\end{align}
where $t$ is the forward time and with
\begin{align} 
\label{eq:Veff}
    V_{\mathrm{eff}}(q)&=\frac{A_t+B_t/\Gamma_t}{2}\,q^2-B_t\log\cosh\left(\frac{\alpha(t)\sqrt{d}}{\Gamma_t}\,q\right)\\
    &=\frac{A_t+B_t/\Gamma_t}{2}\,q^2-B_t\log\cosh\left(\sqrt{\Lambda_t/\Gamma_t}\, q \right)
\end{align} 
where $A_t=\dot\alpha(t)/\alpha(t)$ and $B_t=\big(\dot\beta(t)-A_t\beta(t)\big)\beta(t)$.
We realize that $\sqrt{\Lambda_t/\Gamma_t}$ is the scaling variable controlling the argument of the second term, and that $\Lambda_t$ alone controls it wherever $\Gamma_t=O_d(1)$. Expanding the potential in low and high SNR limits, it reads, close to the origin $q=0$,
\begin{align}
\label{eq:Veff_regimes}
    V_{\mathrm{eff}}(q)\simeq
    \begin{cases}
    \dfrac{1}{2}\left(A_t+\dfrac{B_t}{\Gamma_t}\left(1-\Lambda_t\right)\right)q^2, & \Lambda_t\ll1,\\[12pt]
    \dfrac{1}{2}\left(A_t+\dfrac{B_t}{\Gamma_t}\right)q^2-B_t\,\dfrac{\alpha(t)\sqrt{d}}{\Gamma_t}\,\lvert q\rvert, & \Lambda_t\gg1.
    \end{cases}
\end{align}    
For $\Lambda_t\ll1$ the potential is purely quadratic and the trajectory is not driven towards either mode. For $\Lambda_t\gg1$ it acquires a linear cusp at the origin which makes $q=0$ unstable: the trajectory commits to one of the two branches $q\gtrless0$ and is carried towards the corresponding mode, reaching $q=\pm\sqrt{d}$ at the end of generation.

The transition between these two regimes occurs at some $\Lambda_t=O(1)$, which defines the speciation time $t_s$. We adopt the convention $\Lambda_{t_s}=1$\footnote{Any other choice of $O(1)$ constant shifts $t_s$ only at subleading order in $d$.}. This yields
\begin{align}
\label{eq:speciation_time}
    \frac{\alpha^2(t_s)\,d}{\alpha^2(t_s)\sigma^2+\beta^2(t_s)}=1
    \qquad\Longleftrightarrow\qquad
    \frac{\alpha^2(t_s)}{\beta^2(t_s)}=\frac{1}{d-\sigma^2}\underset{d\gg1}{\sim}\frac1d.
\end{align}
This generalizes the analysis of \citet{biroli_2024} to arbitrary stochastic interpolants. 

\paragraph*{Specific speciations times.} Let us specialize this criterion for the two most widespread SI frameworks, namely Diffusion and Flow Matching. 

\begin{itemize}
\item For Diffusion, $\alpha(t)=e^{-t}$ and $\beta(t)=\sqrt{1-e^{-2t}}$, so that $\Gamma_t=1-e^{-2t}(1-\sigma^2)$ and
\begin{align}
\label{eq:snr_vp}
    \Lambda^{\mathrm{DM}}_t=\frac{e^{-2t}d}{1-e^{-2t}(1-\sigma^2)},
    \qquad
    t_s=\frac12\log\left(d+1-\sigma^2\right)\underset{d\gg1}{\sim}\frac12\log d,
\end{align}
recovering the result of \citet{biroli_2024}. 

\item For the linear interpolant of Flow Matching, $\alpha(t)=1-t$ and $\beta(t)=t$ with $t\in[0,1]$, so that $\Gamma_t=(1-t)^2\sigma^2+t^2$ and
\begin{align}
\label{eq:snr_fm}
    \Lambda^{\mathrm{FM}}_t=\frac{(1-t)^2d}{(1-t)^2\sigma^2+t^2},
    \qquad
    \frac{t_s}{1-t_s}=\sqrt{d-\sigma^2}
    \quad\Longleftrightarrow\quad
    1-t_s=\frac{1}{1+\sqrt{d-\sigma^2}}\underset{d\gg1}{\sim}\frac{1}{\sqrt d}.
\end{align}

\end{itemize}

\subsection{Single time analysis}

\subsubsection{Unbalanced distribution: exact loss and gradient flow}
\label{app:unbalanced_exact}

This section is self-contained. We derive, for the score model actually used in our experiments, the DSM loss and gradient flow exactly as functions of a finite set of summary statistics, valid for any signal-to-noise ratio, and only then specialize to the two regimes of interest by scaling analysis. 

\paragraph*{Setup.} The target is the unbalanced mixture
\begin{align}
\label{eq:target_A}
    P_0=\omega\,\mathcal{N}(\vmu,\sigma^2\vI_d)+(1-\omega)\,\mathcal{N}(-\vmu,\sigma^2\vI_d),
    \qquad \lVert\vmu\rVert^2=d,\qquad \omega,\sigma^2=O_d(1),
\end{align}
and we model the score with a two-layer network with trainable skip connection,
\begin{align}
\label{eq:score_A}
    \vs_{\vtheta=(c,\vw,b)}(\vx,t)=c\,\vx+\vw\tanh\!\left(\vw^T\vx+b\right),
    \qquad c,b\in\mathbb{R},\ \vw\in\mathbb{R}^d.
\end{align}
Note that \eqref{eq:score_A} has no explicit dependence on the schedule $(\alpha,\beta)$: whatever amplitude and argument scale the exact score requires at the noising time $t$, the network must produce through training.
We work at a fixed noising time $t$, writing $\alpha=\alpha(t)$, $\beta=\beta(t)$ and
\begin{align}
\label{eq:Gam_Lam_A}
    \Gamma=\alpha^2\sigma^2+\beta^2,
    \qquad
    \Lambda=\frac{\alpha^2 d}{\Gamma},
    \qquad
    \gamma^2:=\alpha^2 d=\Lambda\Gamma,
\end{align}
where $\Lambda$ is the signal-to-noise ratio and $\gamma$ will turn out to be the natural scale controlling the $\tanh$. Training is gradient flow in training time $\tau$ on the single-time DSM loss,
\begin{align}
\label{eq:loss_def_A}
    \mathcal{L}(\vtheta)=\tfrac1d\,\mathbb{E}_{\vx_0\sim P_0,\ \vxi\sim\mathcal{N}(0,\vI_d)}
    \left[\lVert\beta\,\vs_{\vtheta}(\vx_t,t)+\vxi\rVert^2\right],
    \qquad
    \vx_t=\alpha\vx_0+\beta\vxi,
    \qquad
    \frac{\dd\vtheta}{\dd\tau}=-\nabla_{\vtheta}\mathcal{L}.
\end{align}

\paragraph*{Summary statistics and initialization.} Because \eqref{eq:score_A} carries no $\alpha$, the weight vector must itself grow to the scale set by the exact score. It is therefore convenient to measure $\vw$ in units of $\alpha$ and define
\begin{align}
\label{eq:mq_A}
    m=\frac{\vw^T\vmu}{\alpha\,d},
    \qquad
    q=\frac{\lVert\vw^\perp\rVert^2}{\alpha^2 d},
    \qquad\text{so that}\qquad
    \lVert\vw\rVert^2=\alpha^2 d\left(m^2+q\right),
\end{align}
with $\vw^\perp$ the component of $\vw$ orthogonal to $\vmu$. Equivalently, $m$ and $q$ are the usual overlap and orthogonal norm of the rescaled vector $\vu=\vw/\alpha$. Together with $c$ and $b$, these are the only quantities the loss will depend on.

We initialize $c(0),b(0)=O_d(1)$ and draw $\vw(0)=\vw_0$ uniformly on the unit sphere $\mathbb{S}^{d-1}(1)$, as in our experiments. Then $\vw_0^T\vmu$ has zero mean and variance $\lVert\vmu\rVert^2/d=1$, while $\lVert\vw_0^\perp\rVert^2=1-O_d(1/d)$, so that
\begin{align}
\label{eq:init_A}
    m_0=O_d\!\left(\frac{1}{\alpha d}\right),
    \qquad
    q_0=\frac{1}{\alpha^2 d}\left(1+O_d\!\left(\tfrac1d\right)\right)=\frac{1}{\gamma^2}+O_d\!\left(\tfrac1d\right).
\end{align}
The initial condition thus depends on the noising time through $\alpha$ alone. Everything below establishes the following statement, the long form of Result~\ref{thm:unbalanced} of the main text.

\begin{resultr}[\ref{thm:unbalanced}, long form: training dynamics, unbalanced distribution]
Let $\vw(0)$ be uniform on $\mathbb S^{d-1}(1)$, $c(0),b(0)=O_d(1)$, and let $(c,\vw,b)$ evolve under GF \eqref{eq:grad_flow} for the score model \eqref{eq:score_model_unbalanced} at fixed $t$. As $d\to\infty$:
\begin{enumerate}[label=(\alph*)]
    \item \textbf{Regime~II} ($\Lambda=O_d(d)$). On $\tau=O_d(1)$ times, $c\to c_1^*=-1/(\Gamma+\alpha^2)$ while $(m,q,b)$ remain at initialization; on $\tau=O_d(d)$ times,
    \begin{align}
    \label{eq:RII_thmr_A}
        \dot q=-\frac{4\beta^2}{d}\,q,
        \qquad
        \dot m=-\frac{2\beta^2\Gamma}{d(\Gamma+\alpha^2)}\Big(m-\frac1{\Gamma}\Big),
    \end{align}
    so $(m,q)\to(1/\Gamma,0)$, $c$ is slaved to the value of $(m,q)$ and $b$ stays frozen at its initial value: the network recovers the exact score \eqref{eq:unbalanced} up to its bias, and the imbalance $\omega$ is \textbf{not learned}.
    \item \textbf{Speciation time} ($\Lambda=O_d(1)$). Here $\Gamma\to1$ and $\gamma^2=\Lambda\Gamma=O_d(1)$, with initialization $q(0)=1/\gamma^2=O_d(1)$, $m(0)=O_d(1/\sqrt d)$. On $\tau=O_d(1)$ times $c\to-1/\Gamma$. On $\tau=O_d(d)$ times, i.e.\ in the rescaled training time $\check\tau=\beta^2\tau/d$, the orthogonal component relaxes on its own while, at small $(m,b)$, the overlap and the bias leave the origin jointly,
    \begin{align}
    \label{eq:M_thmr_A}
        \frac{\dd q}{\dd\check\tau}=-4q\,\Psi(q,\gamma)+O(m^2,b^2),
        \qquad
        \frac{\dd}{\dd\check\tau}\begin{pmatrix}m\\b\end{pmatrix}=\vM(q,\gamma)\begin{pmatrix}m\\b\end{pmatrix}+O(m^2,b^2),
    \end{align}
    with the scalar $\Psi$ given in \eqref{eq:lam_Xi_A} and the explicit $2\times2$ matrix $\vM$ in \eqref{eq:coupled_A}.
    The flow converges to $q\to0$, $m\to1/\Gamma$ and $b\to b^*=\tfrac12\log\frac{\omega}{1-\omega}$: the imbalance is learned jointly with the mode direction, on the same $\tau=O_d(d)$ timescale on which Regime~II recovers the direction alone.
\end{enumerate}
\end{resultr}

\paragraph{Detailed dynamics.} 

\paragraph{1. Effective noises.} Decompose the data as $\vx_0=s\vmu+\sigma\vz$, with $s=\pm1$ with probabilities $\omega,1-\omega$ and $\vz\sim\mathcal{N}(0,\vI_d)$, and introduce
\begin{align}
\label{eq:qzqxi_A}
    q_{\vz}=\frac{\vw^T\vz}{\alpha\sqrt{d(m^2+q)}},
    \qquad
    q_{\vxi}=\frac{\vw^T\vxi}{\alpha\sqrt{d(m^2+q)}},
    \qquad
    q_{\vz},q_{\vxi}\overset{\text{iid}}{\sim}\mathcal{N}(0,1),
\end{align}
using $\lVert\vw\rVert^2=\alpha^2d(m^2+q)$ from \eqref{eq:mq_A}. The projections of data and noise onto $\vw$ then read
\begin{align}
\label{eq:proj_A}
    \vw^T\vx_0=\alpha\left(s\,m\,d+\sigma\sqrt{d(m^2+q)}\,q_{\vz}\right),
    \qquad
    \vw^T\vxi=\alpha\sqrt{d(m^2+q)}\,q_{\vxi}.
\end{align}

\paragraph*{2. Effective Loss.} Inserting $\vx_t=\alpha\vx_0+\beta\vxi$ into \eqref{eq:score_A} and collecting terms along $\vx_0$, $\vxi$ and $\vw$,
\begin{align}
\label{eq:regroup_A}
    \beta\,\vs_{\vtheta}(\vx_t,t)+\vxi=\alpha\beta c\,\vx_0+\left(1+\beta^2c\right)\vxi+\beta\,T\,\vw,
    \qquad
    T:=\tanh\!\left(\vw^T\vx_t+b\right).
\end{align}
Squaring, dividing by $d$, and using the concentration of $\tfrac1d\lVert\vx_0\rVert^2\to1+\sigma^2$, $\tfrac1d\lVert\vxi\rVert^2\to1$ and $\tfrac1d\vx_0^T\vxi\to0$, together with \eqref{eq:proj_A}, gives the effective loss
\begin{align}
\label{eq:loss_raw_A}
    \mathcal{L}&=\left(1+\beta^2c\right)^2+\alpha^2\beta^2\left[c^2\left(1+\sigma^2\right)+\left(m^2+q\right)\mathbb{E}\!\left[T^2\right]+2c\,m\,\mathbb{E}\!\left[s\,T\right]\right]\nonumber\\
    &\quad+\frac{2\alpha\beta\sqrt{m^2+q}}{\sqrt d}\Big(\alpha\beta c\,\sigma\,\mathbb{E}\!\left[q_zT\right]+\left(1+\beta^2c\right)\mathbb{E}\!\left[q_{\vxi} T\right]\Big).
\end{align}
While the last line of \eqref{eq:loss_raw_A} carries an explicit $1/\sqrt d$, it is not subleading in $d$. Indeed, $T$ is correlated with $q_{\vz},q_{\vxi}$, and Gaussian integration by parts (Stein's lemma), using $\partial T/\partial q_{\vxi}=\alpha\beta\sqrt{d(m^2+q)}\,(1-T^2)$ and $\partial T/\partial q_{\vz}=\alpha^2\sigma\sqrt{d(m^2+q)}\,(1-T^2)$ from \eqref{eq:proj_A}, gives
\begin{align}
\label{eq:stein_A}
    \mathbb{E}\!\left[q_{\vxi} T\right]=\alpha\beta\sqrt{d(m^2+q)}\;\mathbb{E}\!\left[1-T^2\right],
    \qquad
    \mathbb{E}\!\left[q_{\vz} T\right]=\alpha^2\sigma\sqrt{d(m^2+q)}\;\mathbb{E}\!\left[1-T^2\right].
\end{align}
such that each expectation carries a compensating $\sqrt d$. Substituting into \eqref{eq:loss_raw_A}, the explicit $1/\sqrt d$ cancels exactly and, using $1+\beta^2c+\alpha^2\sigma^2c=1+c\,\Gamma$, the whole line collapses to the $O_d(1)$ contribution 
\begin{align}
\label{eq:collapse_A}
    2\alpha^2\beta^2\left(m^2+q\right)\left(1+c\,\Gamma\right)\mathbb{E}\!\left[1-T^2\right].
\end{align}
yielding
\begin{align}
\label{eq:loss_collapsed_A}
    \mathcal{L}&=\left(1+\beta^2c\right)^2\\
    &\nonumber\quad+\alpha^2\beta^2\left[c^2\left(1+\sigma^2\right)+\left(m^2+q\right)\mathbb{E}\!\left[T^2\right]+2c\,m\,\mathbb{E}\!\left[s\,T\right]\right]\\
    &\nonumber\quad+2\alpha^2\beta^2\left(m^2+q\right)\left(1+c\,\Gamma\right)\mathbb{E}\!\left[1-T^2\right].
\end{align}

\paragraph*{3. $\tanh$ analysis.} From \eqref{eq:proj_A}, $\vw^T\vx_t=\alpha^2 s m d+\alpha\sqrt{d(m^2+q)}\left(\alpha\sigma q_{\vz}+\beta q_{\vxi}\right)$. The two independent Gaussians combine into a single $g\sim\mathcal{N}(0,1)$ with total variance $\alpha^2\sigma^2+\beta^2=\Gamma$, so that, using $\alpha^2d=\gamma^2$,
\begin{align}
\label{eq:U_A}
    \vw^T\vx_t+b=s\,\gamma^2 m+\gamma\sqrt{\Gamma\left(m^2+q\right)}\;g+b.
\end{align}
Averaging over $s$ with probabilities $\omega,1-\omega$, and flipping $g\to-g$ in the $s=-1$ branch, the two expectations left in \eqref{eq:loss_raw_A} reduce to one-dimensional Gaussian integrals,
\begin{align}
\label{eq:hk_A}
    &h:=\mathbb{E}\!\left[s\,T\right]=\omega\,I_1(b)+(1-\omega)\,I_1(-b),
    \qquad
    k:=\mathbb{E}\!\left[T^2\right]=\omega\,I_2(b)+(1-\omega)\,I_2(-b),\\
    &I_1(b)=\mathbb{E}_g\!\left[\tanh\!\left(\gamma^2m+\gamma\sqrt{\Gamma(m^2+q)}\,g+b\right)\right],
    \qquad
    I_2(b)=\mathbb{E}_g\!\left[\tanh^2\!\left(\gamma^2m+\gamma\sqrt{\Gamma(m^2+q)}\,g+b\right)\right].\nonumber
\end{align}

\paragraph*{4. Exact loss.} Collecting \eqref{eq:loss_raw_A}, \eqref{eq:collapse_A} and \eqref{eq:hk_A},
\begin{align}
\label{eq:loss_exact_A}
    \mathcal{L}(c,m,q,b)=\left(1+\beta^2c\right)^2+\alpha^2\beta^2\Big[c^2\left(1+\sigma^2\right)
    +\left(m^2+q\right)\Big(k+2\left(1+c\,\Gamma\right)\left(1-k\right)\Big)+2c\,m\,h\Big],
\end{align}
exactly, for any $\Lambda$, $\omega$ and $b$, with $h,k$ given by \eqref{eq:hk_A}.

\paragraph*{5. Exact gradient flow.} The flow acts on $\vw\in\mathbb{R}^d$, while \eqref{eq:loss_exact_A} depends on it only through $m$ and $q$. From \eqref{eq:mq_A},
\begin{align}
\label{eq:chain_A}
    \nabla_{\vw}m=\frac{\vmu}{\alpha d},
    \qquad
    \nabla_{\vw}q=\frac{2\vw^\perp}{\alpha^2 d},
\end{align}
so that, projecting $\dd\vw/\dd\tau=-\nabla_{\vw}\mathcal{L}$ onto $\vmu$ and onto $\vw^\perp$ and using $\vmu^T\vw^\perp=0$, $\lVert\vmu\rVert^2=d$ and $\lVert\vw^\perp\rVert^2=\alpha^2dq$,
\begin{align}
\label{eq:flow_A}
    \frac{\dd c}{\dd\tau}=-\frac{\partial\mathcal{L}}{\partial c},
    \qquad
    \frac{\dd m}{\dd\tau}=-\frac{1}{\gamma^2}\frac{\partial\mathcal{L}}{\partial m},
    \qquad
    \frac{\dd q}{\dd\tau}=-\frac{4q}{\gamma^2}\frac{\partial\mathcal{L}}{\partial q},
    \qquad
    \frac{\dd b}{\dd\tau}=-\frac{\partial\mathcal{L}}{\partial b}.
\end{align}
The vector parameters $m,q$ inherit a factor $1/\gamma^2=1/(\alpha^2d)$ from the change of variables that the scalars $c,b$ do not. This asymmetry, and the fact that $\gamma^2$ is $O_d(d)$ in one regime and $O_d(1)$ in the other, is what drives everything below.
Equations \eqref{eq:loss_exact_A}--\eqref{eq:flow_A} are the exact reduction of the problem. The two regimes are now obtained by scaling analysis, i.e.\ by evaluating $h,k$ from \eqref{eq:hk_A} in the large $d$ limit.
 
\paragraph*{Regime II: $\Lambda=O_d(d)$, i.e.\ $\alpha=O_d(1)$.} Here $\gamma^2=\alpha^2d=O_d(d)$, so as soon as $m^2+q=O_d(1)$ the Gaussian term in \eqref{eq:U_A} has standard deviation $\gamma\sqrt{\Gamma(m^2+q)}=O_d(\sqrt d)$: the $\tanh$ argument diverges for almost every $g$ and $T\to\sign$. Consequently $k\to1$, and the $O_d(1)$ bias $b$ is swamped, so $h\to\mathbb{E}[\sign(\nu+g)]=\mathrm{erf}(\nu/\sqrt2)$ with
\begin{align}
\label{eq:nu_A}
    \nu=\sqrt{\Lambda}\,\frac{m}{\sqrt{m^2+q}}.
\end{align}
Crucially $1-k\to0$ exponentially in $\gamma$ (a Gaussian tail), so the term \eqref{eq:collapse_A} is genuinely negligible here, and $h$ is independent of $b$. Since the imbalance $\omega$ enters the exact score \eqref{eq:unbalanced} only through the offset $\tfrac12\log\tfrac{\omega}{1-\omega}$, and $\partial\mathcal{L}/\partial b\to0$, the bias never moves in this regime. The loss reduces to
\begin{align}
\label{eq:loss_RII_A}
    \mathcal{L}(c,m,q)=\left(1+\beta^2c\right)^2+\alpha^2\beta^2\left[c^2\left(1+\sigma^2\right)+m^2+q+2c\,m\,\mathrm{erf}\!\left(\frac{\nu}{\sqrt2}\right)\right],
\end{align}
and \eqref{eq:flow_A} closes on $(c,m,q)$, with all the $\alpha^2$ prefactors cancelling against $1/\gamma^2$ in the $m,q$ equations:
\begin{align}
\label{eq:ode_RII_A}
    \frac{\dd c}{\dd\tau}&=-2\beta^2\left(1+\beta^2c\right)-2\alpha^2\beta^2\left[c\left(1+\sigma^2\right)+m\,\mathrm{erf}\!\left(\frac{\nu}{\sqrt2}\right)\right],\\
    \frac{\dd m}{\dd\tau}&=-\frac{2\beta^2}{d}\left[m+c\,\mathrm{erf}\!\left(\frac{\nu}{\sqrt2}\right)+\sqrt{\frac2\pi}\,\frac{c\,q\,m\sqrt{\Lambda}}{\left(m^2+q\right)^{3/2}}\,e^{-\nu^2/2}\right],\\
    \frac{\dd q}{\dd\tau}&=-\frac{4\beta^2}{d}\,q\left[1-\sqrt{\frac2\pi}\,\frac{c\,m^2\sqrt{\Lambda}}{\left(m^2+q\right)^{3/2}}\,e^{-\nu^2/2}\right].
\end{align}
The right-hand sides of the $m$ and $q$ equations are $O_d(1/d)$ while that of $c$ is $O_d(1)$: the dynamics splits into two phases.

Phase 1, $\tau=O_d(1)$. Only $c$ moves. By \eqref{eq:init_A}, $m_0,q_0=O_d(1/d)$, so $m\,\mathrm{erf}(\nu/\sqrt2)$ is negligible and
\begin{align}
\label{eq:c1_A}
    \frac{\dd c}{\dd\tau}=-2\beta^2\left(\Gamma+\alpha^2\right)\left(c-c^*_1\right),
    \qquad
    c^*_1=-\frac{1}{\Gamma+\alpha^2}=-\frac{d}{\mathbb{E}\lVert\vx_t\rVert^2},
\end{align}
so $c$ relaxes exponentially to the value that best fits the isotropic Gaussian approximation of $P_t$, while $(m,q,b)$ sit at initialization.

Phase 2, $\tau=O_d(d)$. Now $m,q$ evolve while $c$, being $d$ times faster, is slaved through $\partial_c\mathcal{L}=0$, i.e.\ $c^*=-(1+\alpha^2m\,\mathrm{erf}(\nu/\sqrt2))/(\Gamma+\alpha^2)$. Note from \eqref{eq:init_A} and \eqref{eq:nu_A} that $\nu=O_d(1)$ already at initialization in this regime: the drive $-c\,\mathrm{erf}(\nu/\sqrt2)>0$ is present from the outset and $m$ grows immediately, with no instability threshold to cross. Once $\nu\gg1$ the error function saturates and, taking $m>0$ without loss of generality, the flow linearizes,
\begin{align}
\label{eq:phase2_A}
    \frac{\dd q}{\dd\tau}=-\frac{4\beta^2}{d}\,q,
    \qquad
    \frac{\dd m}{\dd\tau}=-\frac{2\beta^2\Gamma}{d\left(\Gamma+\alpha^2\right)}\left(m-\frac{1}{\Gamma}\right),
\end{align}
so $q\to0$ and $m\to1/\Gamma$ on $\tau=O_d(d)$. At the fixed point $\vw=\alpha\vmu/\Gamma$ and $c=-1/\Gamma$, giving
\begin{align}
\label{eq:score_end_A}
    \vs_{\vtheta}(\vx,t)\longrightarrow-\frac{\vx}{\Gamma}+\frac{\alpha\vmu}{\Gamma}\tanh\!\left(\frac{\alpha\,\vmu^T\vx}{\Gamma}+b_0\right),
\end{align}
which is the exact score \eqref{eq:unbalanced} of the target except for the frozen bias $b_0$ in place of $b^*=\tfrac12\log\tfrac{\omega}{1-\omega}$.

\paragraph*{Speciation time: $\Lambda=O_d(1)$, i.e.\ $\alpha^2=O_d(1/d)$.} Now $\gamma^2=\Lambda\Gamma=O_d(1)$ and the $\tanh$ argument \eqref{eq:U_A} stays $O_d(1)$: no saturation, and the bias is no longer negligible. Two simplifications are available. First, $\alpha^2=O_d(1/d)\to0$ forces $t\to\tmax$, hence $\beta\to1$ and
\begin{align}
\label{eq:Gam_one_A}
    \Gamma=\alpha^2\sigma^2+\beta^2\longrightarrow1,
\end{align}
so we set $\Gamma=1$ and $\gamma^2=\Lambda$ throughout. Second, by \eqref{eq:init_A} the initial condition is now
\begin{align}
\label{eq:init_spec_A}
    q_0=\frac{1}{\gamma^2}=O_d(1),
    \qquad
    m_0=O_d\!\left(\frac{1}{\gamma\sqrt d}\right)=O_d\!\left(\frac{1}{\sqrt d}\right),
\end{align}
so $m$ starts vanishingly small against an $O_d(1)$ value of $q$ - the opposite of Regime~II, where both were $O_d(1/d)$. 
Since $\alpha^2\beta^2=O_d(1/d)$, the loss \eqref{eq:loss_exact_A} is dominated by $(1+\beta^2c)^2$ and $c$ relaxes on $\tau=O_d(1)$ to $c^*=-1/\Gamma+O_d(1/d) = -1+O_d(1/d)$, before $(m,q,b)$ move. Substituting $c=c^*$ makes the $(1+c\Gamma)$ term in \eqref{eq:loss_exact_A} $O_d(1/d)$, leaving $\mathcal{L}=\mathrm{const}+\alpha^2\beta^2\,\mathcal{F}+O_d(1/d^2)$ with
\begin{align}
\label{eq:F_A}
    \mathcal{F}(m,q,b)=\mathbb{E}_{g,s}\Big[\left(m^2+q\right)T^2-2\,s\,m\,T\Big],
    \qquad
    T=\tanh(A),
    \qquad
    A=s\,\gamma^2m+\gamma\sqrt{m^2+q}\;g+b.
\end{align}
Measuring training time in units of $\check\tau=\beta^2\tau/d$, the exact flow \eqref{eq:flow_A} becomes
\begin{align}
\label{eq:flow_spec_A}
    \frac{\dd m}{\dd\check\tau}=-\frac{\partial\mathcal{F}}{\partial m},
    \qquad
    \frac{\dd q}{\dd\check\tau}=-4q\,\frac{\partial\mathcal{F}}{\partial q},
    \qquad
    \frac{\dd b}{\dd\check\tau}=-\gamma^2\,\frac{\partial\mathcal{F}}{\partial b},
\end{align}
where the $\alpha^2$ of the loss has cancelled against the $1/\gamma^2=1/(\alpha^2d)$ of \eqref{eq:flow_A} in the $m,q$ equations, while the $b$ equation retains a factor $\gamma^2=O_d(1)$. All four parameters therefore move at comparable rates in $\check\tau$, i.e.\ $\tau = O_d(d)$.

Linear stability of $m=b=0$. Expanding \eqref{eq:F_A} to second order in $(m,b)$ at fixed $q$, and introducing the Gaussian averages
\begin{align}
\label{eq:Theta_A}
    \Theta_1(q,\gamma)=\mathbb{E}_g\!\left[\tanh^2\!\left(\gamma\sqrt q\,g\right)\right],
    \qquad
    \Theta_2(q,\gamma)=\mathbb{E}_g\!\left[\left(1-\tanh^2\right)\left(1-3\tanh^2\right)\!\left(\gamma\sqrt q\,g\right)\right],
\end{align}
one finds $\partial_m\mathcal{F}=\partial_b\mathcal{F}=0$ at $m=b=0$, both integrands being odd. Writing
\begin{align}
\label{eq:lam_Xi_A}
    &\Psi(q,\gamma)=\Theta_1+\gamma^2q\,\Theta_2,\\
    &\nonumber\mu(q,\gamma)=4\gamma^2\left(1-\Theta_1\right)-2\Theta_1-2\gamma^2q\left(1+\gamma^2\right)\Theta_2,\\
    &\nonumber\Xi(q,\gamma)=\left(1-\Theta_1\right)-\gamma^2q\,\Theta_2,
\end{align}
the zeroth order of the expansion gives the autonomous relaxation of $q$,
\begin{align}
\label{eq:q_alone_A}
    \frac{\dd q}{\dd\check\tau}=-4q\,\Psi(q,\gamma)+O(m^2,b^2),
\end{align}
which decays monotonically from $q_0=1/\gamma^2$, while the first order gives the coupled linear system obeyed by the overlap and the bias at fixed $q$,
\begin{align}
\label{eq:coupled_A}
    \frac{\dd}{\dd\check\tau}\begin{pmatrix}m\\b\end{pmatrix}
    =\vM(q,\gamma)\begin{pmatrix}m\\b\end{pmatrix},
    \qquad
    \vM(q,\gamma)=\begin{pmatrix}
    \mu & 2\left(2\omega-1\right)\Xi\\[2pt]
    2\gamma^2\left(2\omega-1\right)\Xi & -2\gamma^2q\,\Theta_2
    \end{pmatrix}.
\end{align}
The off-diagonal terms are proportional to $2\omega-1$: for balanced data the overlap and the bias decouple entirely, $m$ growing at the rate $\mu(q,\gamma)$ while $b$ decays to $0$. For $\omega\neq1/2$ the two are coupled and leave the origin together, the flow converging to $q=0$, $m=1/\Gamma$ and
\begin{align}
\label{eq:bstar_A}
    b\longrightarrow b^*=\frac12\log\frac{\omega}{1-\omega},
    \qquad\text{i.e.}\qquad
    \rho\longrightarrow\omega .
\end{align}
Contrary to Regime~II, where the bias stays frozen at $b_0$ and only the direction is recovered, training around the speciation time learns the mode direction and the imbalance, jointly and on the same $\tau=O_d(d)$ timescale.

\subsubsection{GMM with four modes: exact loss and gradient flow}
\label{app:hierarchical_exact}

This section is self-contained, and mirrors \S\ref{app:unbalanced_exact} for the hierarchical target, using the score model of the main text, \eqref{eq:score_model_4modes}. 

\paragraph*{Setup.} The target is the block-structured quadrimodal mixture
\begin{align}
\label{eq:target_B}
    P_0(\vx)=\frac14\sum_{s_1=\pm1}\sum_{s_2=\pm1}\mathcal{N}\!\left(s_1\vmu_1+s_2\vmu_2,\,\sigma^2\vI_d\right),
\end{align}
where $\vmu_1$ is supported on the first $\kappa d$ coordinates and $\vmu_2$ on the remaining $(1-\kappa)d$. We write $\kappa_1=\kappa$, $\kappa_2=1-\kappa$, both $O_d(1)$, and normalize $\lVert\vmu_i\rVert^2=\kappa_id$ so that every coordinate of the means is $O_d(1)$. The score model is
\begin{align}
\label{eq:score_B}
    \vs_{\vtheta=(c,\vw_1,\vw_2)}(\vx,t)=c\,\vx+\vw_1\tanh\!\left(\vw_1^T\vx\right)+\vw_2\tanh\!\left(\vw_2^T\vx\right),
\end{align}
with $\vw_i$ supported on block $i$, so that $\vw_1^T\vw_2=0$ and $\vw_i^T\vmu_j=0$ for $i\ne j$. As in \S\ref{app:unbalanced_exact} we work at fixed noising time $t$, write $\alpha=\alpha(t),\beta=\beta(t)$, $\Gamma=\alpha^2\sigma^2+\beta^2$, $\Lambda=\alpha^2d/\Gamma$, and train by gradient flow on the single-time DSM loss $\mathcal{L}=\tfrac1d\mathbb{E}\lVert\beta\vs_{\vtheta}(\vx_t,t)+\vxi\rVert^2$ with $\vx_t=\alpha\vx_0+\beta\vxi$. Note that \eqref{eq:score_B} carries no bias: each block of \eqref{eq:target_B} is balanced, so there is no mode weight to learn, and the structural parameter is $\kappa$ instead.

\paragraph*{Summary statistics, block SNR, initialization.} Per block we define, as in \eqref{eq:order_params},
\begin{align}
\label{eq:mq_B}
    m_i=\frac{\vw_i^T\vmu_i}{\alpha\,\kappa_id},
    \qquad
    q_i=\frac{\lVert\vw_i^\perp\rVert^2}{\alpha^2\kappa_id},
    \qquad\text{so that}\qquad
    \lVert\vw_i\rVert^2=\alpha^2\kappa_id\left(m_i^2+q_i\right),
\end{align}
with $\vw_i^\perp$ the component of $\vw_i$ orthogonal to $\vmu_i$ inside block $i$. Because block $i$ has $\kappa_id$ coordinates and $\lVert\vmu_i\rVert^2=\kappa_id$, the natural signal-to-noise ratio and $\tanh$ scale of block $i$ are those of \S\ref{app:unbalanced_exact} with $d\to\kappa_id$:
\begin{align}
\label{eq:snr_B}
    \Lambda^{(i)}=\frac{\alpha^2\lVert\vmu_i\rVert^2}{\Gamma}=\kappa_i\Lambda,
    \qquad
    \gamma_i^2:=\alpha^2\kappa_id=\kappa_i\Lambda\Gamma=\kappa_i\gamma^2 .
\end{align}
The block asymmetry $\kappa$ enters only through this rescaling of the SNR. Initializing each $\vw_i(0)$ uniformly on the unit sphere of its own block gives, exactly as in \eqref{eq:init_A},
\begin{align}
\label{eq:init_B}
    m_i(0)=O_d\!\left(\frac{1}{\alpha\kappa_id}\right),
    \qquad
    q_i(0)=\frac{1}{\gamma_i^2}+O_d\!\left(\tfrac1d\right),
    \qquad\text{so that}\qquad
    \gamma_i\sqrt{q_i(0)}=1
\end{align}
for both blocks, whatever $\kappa$: the two blocks start at the same point in the rescaled variable that controls the $\tanh$, and differ only through $\gamma_i$. Together with (\S\ref{app:fixednorm}), everything below establishes the following statement, the long form of Result~\ref{thm:hierarchical} of the main text.\\

\begin{resultr}[\ref{thm:hierarchical}, long form: training dynamics, hierarchical distribution]
Let $\vw_1(0),\vw_2(0)$ be independent and uniform on the unit sphere of their respective $\kappa d$ (resp. $(1-\kappa)d$) dimensional blocks, $c(0)=O_d(1)$, and let $(c,\vw_1,\vw_2)$ evolve under GF \eqref{eq:grad_flow} for the score model \eqref{eq:score_model_4modes} at fixed $t$. 
Define $\kappa_1=\kappa, \kappa_2=1-\kappa$, each pair $(m_i,q_i)$ obeys the dynamics of Result~\ref{thm:unbalanced} at $\omega=1/2$, $b=0$ and rescaled SNR $\Lambda\to\kappa_i\Lambda$, the two blocks being coupled only through the shared skip connection $c$. As $d\to\infty$:
\begin{enumerate}[label=(\alph*)]
    \item \textbf{Regime~II.} On $\tau=O_d(1)$ times only $c$ moves, relaxing to $c^*_1=-1/(\Gamma+\alpha^2)$ while both $(m_i,q_i)$ stay at initialization. On $\tau=O_d(d)$ times $c$ is slaved to the overlaps, and $(m_1,q_1)$, $(m_2,q_2)$ both converge to $(1/\Gamma,0)$, independently of $\kappa_i$: the two mode directions $\vmu_1,\vmu_2$ are recovered together, on the same timescale, regardless of the block asymmetry.
    \item \textbf{Speciation time} ($\Lambda=O_d(1)$). Here $\Gamma\to1$, each block carries its own SNR $\gamma_i^2=\kappa_i\Lambda\Gamma$ and starts from $q_i(0)=1/\gamma_i^2$. On $\tau=O_d(1)$ times $c\to-1/\Gamma$ and stays there for the remainder of the dynamics; with $c$ thus frozen the two blocks decouple entirely, and since $\omega=1/2$ makes $\vM(q_i,\gamma_i)$ diagonal, each pair obeys the closed autonomous system
    \begin{align}
    \label{eq:joint_mq_thmr_B}
        \frac{\dd q_i}{\dd\check\tau}=-4q_i\,\Psi\!\left(q_i,\gamma_i\right),
        \qquad
        \frac{\dd m_i}{\dd\check\tau}=\mu\!\left(q_i,\gamma_i\right)m_i+O(m_i^3),
    \end{align}
    with $\Psi$ and the growth rate $\mu$ as given in \eqref{eq:lam_Xi_A}.
    The orthogonal component $q_i$ decreases monotonically on the same timescale as $m_i$, and $\mu$ is a strictly decreasing function of $q_i$: it is the decay of $q_i$ that drives the overlap unstable.
    In the $q_i\to 0$ limit, $\mu(q_i, \gamma_i)\sim 4 \gamma_i^2 \equiv 4\kappa_i\Lambda$: the rate of block $i$ is proportional to $\kappa_i$. The two directions are learned on different timescales - for $\kappa$ small the dominant direction $\vmu_2$ is learned on $\tau=O_d(d)$, while the subdominant $\vmu_1$ lags by a factor $1/\kappa$, on $\tau=O_d(d/\kappa)$.
    \item \textbf{Heuristic growth rate.} The system \eqref{eq:joint_mq_thmr_B} is two-dimensional, so at finite $\gamma_i$ the rate $\mu(q_i,\gamma_i)$ drifts as $q_i$ relaxes. Assuming instead that $\lVert\vw_i\rVert$ is initialized and kept at the norm $\alpha\lVert\vmu_i\rVert$ of the target direction, i.e.\ $m_i^2+q_i=1$, eliminates $q_i$ from \eqref{eq:joint_mq_thmr_B} and reduces the dynamics to a single ODE $\dd m_i/\dd\check\tau=\lambda(\gamma_i)\,m_i+O(m_i^3)$, with the closed-form rate \eqref{eq:rate_C} of \S\ref{app:fixednorm},
    \begin{align}
    \label{eq:mu_thmr_B}
        \lambda(\gamma)=\mu(1,\gamma)+2\,\Psi(1,\gamma).
    \end{align}
    Note that this constraint is not exactly met by the initialization above, which gives $q_i(0)=1/\gamma_i^2$ rather than $q_i(0)=1$; for $\kappa_i=O_d(1)$, however, the two differ only by the $O_d(1)$ factor $\gamma_i^2=\kappa_i\Lambda\Gamma$, so both start the dynamics at the same order in $d$. The reduction moreover recovers (b) exactly as $\kappa\to0$, since $\lambda(\gamma)=4\gamma^2-6\gamma^4+O(\gamma^6)$. Away from that limit it is a heuristic, but a closed-form one, and we use $\tau_i\propto1/\lambda(\gamma_i)$ as the reference timescale against which the measured dynamics are rescaled in Fig.~\ref{fig:exscore-kappa}.
\end{enumerate}
\end{resultr}

\paragraph{Detailed dynamics.} 

\paragraph*{1. Exact loss.} Decompose $\vx_0=s_1\vmu_1+s_2\vmu_2+\sigma\vz$ with $s_1,s_2=\pm1$ independent and uniform, $\vz\sim\mathcal{N}(0,\vI_d)$, and set $q_{\vz_i}=\vw_i^T\vz/(\alpha\sqrt{\kappa_id(m_i^2+q_i)})$ and $q_{\vxi_i}=\vw_i^T\vxi/(\alpha\sqrt{\kappa_id(m_i^2+q_i)})$; the four are i.i.d.\ standard Gaussians because $\vw_1^T\vw_2=0$. Since $\vw_i^T\vmu_j=0$ for $i\neq j$, block $i$ only ever sees $s_i$, and
\begin{align}
\label{eq:proj_B}
    \vw_i^T\vx_0=\alpha\left(s_i\,m_i\,\kappa_id+\sigma\sqrt{\kappa_id(m_i^2+q_i)}\,q_{\vz_i}\right),
    \qquad
    \vw_i^T\vxi=\alpha\sqrt{\kappa_id(m_i^2+q_i)}\,q_{\vxi_i}.
\end{align}
Inserting $\vx_t=\alpha\vx_0+\beta\vxi$ into \eqref{eq:score_B} and regrouping,
\begin{align}
\label{eq:regroup_B}
    \beta\vs_{\vtheta}(\vx_t,t)+\vxi=\alpha\beta c\,\vx_0+\left(1+\beta^2c\right)\vxi+\beta\sum_{i=1,2}T_i\,\vw_i,
    \qquad
    T_i:=\tanh\!\left(\vw_i^T\vx_t\right).
\end{align}
Squaring and dividing by $d$: the cross term between the two blocks vanishes by $\vw_1^T\vw_2=0$, $\tfrac1d\mathbb{E}\lVert\vx_0\rVert^2\to1+\sigma^2$ since $\lVert\vmu_1\rVert^2+\lVert\vmu_2\rVert^2=d$, and the terms in $q_{\vz_i},q_{\vxi_i}$ carry an explicit $1/\sqrt d$ which is again compensated by Stein's lemma,
\begin{align}
\label{eq:stein_B}
    \mathbb{E}\!\left[q_{\vxi_i}T_i\right]=\alpha\beta\sqrt{\kappa_id(m_i^2+q_i)}\,\mathbb{E}\!\left[1-T_i^2\right],
    \quad
    \mathbb{E}\!\left[q_{\vz_i}T_i\right]=\alpha^2\sigma\sqrt{\kappa_id(m_i^2+q_i)}\,\mathbb{E}\!\left[1-T_i^2\right],
\end{align}
the two recombining through $1+\beta^2c+\alpha^2\sigma^2c=1+c\,\Gamma$. Finally, from \eqref{eq:proj_B} and $\alpha\sqrt{\kappa_id}=\gamma_i$, the argument of the $i$-th $\tanh$ is
\begin{align}
\label{eq:U_B}
    \vw_i^T\vx_t=s_i\,\gamma_i^2m_i+\gamma_i\sqrt{\Gamma\left(m_i^2+q_i\right)}\;g_i,
    \qquad g_i\sim\mathcal{N}(0,1),
\end{align}
and, $s_i$ being uniform, averaging over it with $g_i\to s_ig_i$ removes $s_i$ altogether. Collecting,
\begin{align}
\label{eq:loss_exact_B}
    \mathcal{L}(c,m_1,q_1,m_2,q_2)=\left(1+\beta^2c\right)^2+\alpha^2\beta^2\left[c^2\left(1+\sigma^2\right)
    +\sum_{i=1,2}\kappa_i\,\mathcal{G}_i\right],
\end{align}
\begin{align}
\label{eq:Gi_B}
    \mathcal{G}_i=\left(m_i^2+q_i\right)\Big(k_i+2\left(1+c\,\Gamma\right)\left(1-k_i\right)\Big)+2c\,m_i\,h_i,
\end{align}
with the one-dimensional Gaussian averages
\begin{align}
\label{eq:hk_B}
    h_i=\mathbb{E}_g\!\left[\tanh A_i\right],
    \quad
    k_i=\mathbb{E}_g\!\left[\tanh^2 A_i\right],
    \quad
    A_i=\gamma_i^2m_i+\gamma_i\sqrt{\Gamma\left(m_i^2+q_i\right)}\,g .
\end{align}
Comparing with \eqref{eq:loss_exact_A}: each $\mathcal{G}_i$ is exactly the bracket of the unbalanced problem at $\omega=1/2$ and $b=0$, with $\gamma\to\gamma_i$. The hierarchical target is therefore two independent copies of the single-mode problem at rescaled SNR $\kappa_i\Lambda$, weighted by $\kappa_i$ and coupled only through the shared skip connection $c$.

\paragraph*{2. Exact gradient flow.} With $\nabla_{\vw_i}m_i=\vmu_i/(\alpha\kappa_id)$ and $\nabla_{\vw_i}q_i=2\vw_i^\perp/(\alpha^2\kappa_id)$, projecting $\dd\vw_i/\dd\tau=-\nabla_{\vw_i}\mathcal{L}$ as in \eqref{eq:chain_A} gives
\begin{align}
\label{eq:flow_B}
    \frac{\dd c}{\dd\tau}=-\frac{\partial\mathcal{L}}{\partial c},
    \qquad
    \frac{\dd m_i}{\dd\tau}=-\frac{1}{\gamma_i^2}\frac{\partial\mathcal{L}}{\partial m_i}
    =-\frac{\alpha^2\beta^2\kappa_i}{\gamma_i^2}\frac{\partial\mathcal{G}_i}{\partial m_i},
    \qquad
    \frac{\dd q_i}{\dd\tau}=-\frac{4q_i}{\gamma_i^2}\frac{\partial\mathcal{L}}{\partial q_i}
    =-\frac{4q_i\,\alpha^2\beta^2\kappa_i}{\gamma_i^2}\frac{\partial\mathcal{G}_i}{\partial q_i}.
\end{align}
Using $\gamma_i^2=\alpha^2\kappa_id$ and the rescaled training time $\check\tau=\beta^2\tau/d$ yields 
\begin{align}
\label{eq:flow_resc_B}
    \frac{\dd m_i}{\dd\check\tau}=-\frac{\partial\mathcal{G}_i}{\partial m_i},
    \qquad
    \frac{\dd q_i}{\dd\check\tau}=-4q_i\,\frac{\partial\mathcal{G}_i}{\partial q_i}.
\end{align}
The block asymmetry acts on the dynamics exclusively through $\gamma_i^2=\kappa_i\Lambda\Gamma$ inside $\mathcal{G}_i$, i.e.\ purely as a rescaling of the block SNR.

\paragraph*{Regime II: $\Lambda=O_d(d)$.} Then $\gamma_i^2=\kappa_i\Lambda\Gamma=O_d(d)$ for both blocks, the argument \eqref{eq:U_B} diverges, $T_i\to\sign$, $k_i\to1$ with $1-k_i$ exponentially small, and $h_i\to\mathrm{erf}(\nu_i/\sqrt2)$ with $\nu_i=\sqrt{\kappa_i\Lambda}\,m_i/\sqrt{m_i^2+q_i}$. The loss becomes
\begin{align}
\label{eq:loss_RII_B}
    \mathcal{L}=\left(1+\beta^2c\right)^2+\alpha^2\beta^2\left[c^2\left(1+\sigma^2\right)+\sum_{i=1,2}\kappa_i\left(m_i^2+q_i+2c\,m_i\,\mathrm{erf}\!\left(\frac{\nu_i}{\sqrt2}\right)\right)\right],
\end{align}
and \eqref{eq:flow_B} gives
\begin{align}
\label{eq:ode_RII_B}
    \frac{\dd c}{\dd\tau}&=-2\beta^2\left(1+\beta^2c\right)-2\alpha^2\beta^2\left[c\left(1+\sigma^2\right)+\sum_{i}\kappa_i\,m_i\,\mathrm{erf}\!\left(\frac{\nu_i}{\sqrt2}\right)\right],\\
    \frac{\dd m_i}{\dd\tau}&=-\frac{2\beta^2}{d}\left[m_i+c\,\mathrm{erf}\!\left(\frac{\nu_i}{\sqrt2}\right)+\sqrt{\frac2\pi}\,\frac{c\,q_i\,m_i\sqrt{\kappa_i\Lambda}}{\left(m_i^2+q_i\right)^{3/2}}e^{-\nu_i^2/2}\right],\\
    \frac{\dd q_i}{\dd\tau}&=-\frac{4\beta^2}{d}\,q_i\left[1-\sqrt{\frac2\pi}\,\frac{c\,m_i^2\sqrt{\kappa_i\Lambda}}{\left(m_i^2+q_i\right)^{3/2}}e^{-\nu_i^2/2}\right].
\end{align}
Again the $c$ equation is $O_d(1)$ and the others $O_d(1/d)$, giving two phases. On $\tau=O_d(1)$, with $m_i,q_i=O_d(1/d)$ at initialization, $c$ relaxes to
\begin{align}
\label{eq:c1_B}
    c^*_1=-\frac{1}{\Gamma+\alpha^2}
\end{align}
while both $(m_i,q_i)$ stay put. On $\tau=O_d(d)$, $c$ is slaved through $\partial_c\mathcal{L}=0$, i.e.\ $c^*=-\left(1+\alpha^2\sum_j\kappa_jm_j\,\mathrm{erf}(\nu_j/\sqrt2)\right)/(\Gamma+\alpha^2)$ and linearizing
\begin{align}
\label{eq:phase2_B}
    \frac{\dd q_i}{\dd\tau}=-\frac{4\beta^2}{d}\,q_i,
    \qquad
    \frac{\dd m_i}{\dd\tau}=-\frac{2\beta^2}{d}\left[m_i-\frac{1+\alpha^2\sum_j\kappa_jm_j}{\Gamma+\alpha^2}\right].
\end{align}
Both blocks obey the same dynamics and converge together. Using $\sum_j\kappa_j=1$, the fixed point is $m_1=m_2=1/\Gamma$, $q_i=0$, $c=-1/\Gamma$, whence $\vw_i=\alpha\vmu_i/\Gamma$ and
\begin{align}
\label{eq:score_end_B}
    \vs_{\vtheta}(\vx,t)\longrightarrow-\frac{\vx}{\Gamma}+\sum_{i=1,2}\frac{\alpha\vmu_i}{\Gamma}\tanh\!\left(\frac{\alpha\,\vmu_i^T\vx}{\Gamma}\right),
\end{align}
exactly the target score \eqref{eq:hierarchical}. In Regime~II both mode directions are recovered on the same $\tau=O_d(d)$ timescale, independently of $\kappa$: training at high SNR is blind to the block asymmetry.

\paragraph*{Speciation time: $\Lambda=O_d(1)$.} Now $\gamma_i^2=\kappa_i\Lambda\Gamma=O_d(1)$, and as in \S\ref{app:unbalanced_exact} $\alpha^2=O_d(1/d)$ forces $\Gamma\to1$, so $\gamma_i^2\to\kappa_i\Lambda$. The skip connection relaxes on $\tau=O_d(1)$ to $c^*=-1 + O_d(1/d)$; substituting it kills the $(1+c\Gamma)$ term in \eqref{eq:Gi_B} and leaves $\mathcal{G}_i=\mathcal{F}_i+O_d(1/d)$ with
\begin{align}
\label{eq:F_B}
    \mathcal{F}_i(m_i,q_i)=\mathbb{E}_g\!\left[\left(m_i^2+q_i\right)\tanh^2 A_i-2\,m_i\tanh A_i\right],
    \qquad
    A_i=\gamma_i^2m_i+\gamma_i\sqrt{m_i^2+q_i}\;g .
\end{align}
With $c$ frozen, the two blocks are now completely decoupled: by \eqref{eq:flow_resc_B} each pair $(m_i,q_i)$ follows its own autonomous flow, on the timescale
\begin{align}
\label{eq:tau_B}
    \check\tau=O_d(1)\qquad\Longleftrightarrow\qquad \tau=O_d(d).
\end{align}
Expanding \eqref{eq:F_B} in $m_i$ at fixed $q_i$ (the $b=0$, $\omega=1/2$), and writing $\Theta_{1,2}$ for the Gaussian averages \eqref{eq:Theta_A} evaluated at $(q_i,\gamma_i)$,
\begin{align}
\label{eq:lin_B}
    \frac{\dd q_i}{\dd\check\tau}=-4q_i\,\Psi\!\left(q_i,\gamma_i\right)+O(m_i^2),
    \qquad
    \frac{\dd m_i}{\dd\check\tau}=\mu\!\left(q_i,\gamma_i\right)m_i+O(m_i^3),
\end{align}
with $\mu$ as in \eqref{eq:lam_Xi_A}.
Since, $\Theta_1\to0$ and $\Theta_2\to1$ as $q_i\to0$, $\mu$ increases monotonically along the flow towards
\begin{align}
\label{eq:lam_lim_B}
    \mu\!\left(q_i\to0,\gamma_i\right)=4\gamma_i^2=4\,\kappa_i\Lambda\Gamma\;\xrightarrow[\ \Gamma\to1\ ]{}\;4\,\kappa_i\Lambda :
\end{align}
the asymptotic growth rate of block $i$ is linear in $\kappa_i$. We expect the escape time of $m_i$ to therefore scale as $1/\kappa_i$ in $\check\tau$, i.e.
\begin{align}
\label{eq:tau_kappa_B}
    \tau_i=O_d\!\left(\frac{d}{\kappa_i}\right).
\end{align}
Taking $\kappa$ small, so that $\kappa_1=\kappa\ll1$ and $\kappa_2=1-\kappa\simeq1$, the dominant direction $\vmu_2$ is learned on $\tau=O_d(d)$ while the subdominant direction $\vmu_1$ lags by a factor $1/\kappa$, on $\tau=O_d(d/\kappa)$. Once escaped, each block converges to $q_i=0$, $m_i=1/\Gamma$, recovering \eqref{eq:score_end_B}.

\subsubsection{Fixed-norm ansatz: a one-dimensional reduction at the speciation time}
\label{app:fixednorm}

The analyses of \S\ref{app:unbalanced_exact} and \S\ref{app:hierarchical_exact} involve two coupled order parameters, the overlap $m$ and the orthogonal norm $q$, whose joint relaxation makes the growth rate of $m$ depend on the instantaneous value of $q$. We show here that constraining the weight vector to a fixed norm removes $q$ entirely, reduces the dynamics to a single scalar ODE, and yields a closed-form growth rate at small overlap $m$.

\paragraph*{Fixed norm  constraint.} We keep the notation of \S\ref{app:unbalanced_exact}: score model \eqref{eq:score_A}, order parameters \eqref{eq:mq_A}, and $\gamma^2=\alpha^2d=\Lambda\Gamma$. We now impose that $\vw$ retain throughout training the norm of the target direction,
\begin{align}
\label{eq:constraint_C}
    \lVert\vw\rVert=\alpha\sqrt d
    \qquad\Longleftrightarrow\qquad
    m^2+q=1,
\end{align}
using $\lVert\vw\rVert^2=\alpha^2d(m^2+q)$. This is the natural scale: the exact score \eqref{eq:unbalanced} is reproduced at $\vw^*=\alpha\vmu/\Gamma$, of norm $\alpha\sqrt d/\Gamma$.
We place ourselves at the speciation time, $\Lambda=O_d(1)$, and set $\Gamma=1$ throughout, so that $\gamma^2=\Lambda$.

\paragraph*{The loss loses its dependence on $q$.} The whole $(m,q)$ dependence of the loss enters through the law \eqref{eq:U_A} of the $\tanh$ argument, whose mean is $\propto m$ and whose standard deviation is $\gamma\sqrt{\Gamma(m^2+q)}$. Under \eqref{eq:constraint_C} the latter is constant,
\begin{align}
\label{eq:U_C}
    \vw^T\vx_t+b=s\,\gamma^2m+\gamma\,g+b,
    \qquad g\sim\mathcal{N}(0,1),
\end{align}
so the constraint freezes the width of the Gaussian and lets only its mean move with $m$. Writing $\mathcal{F}$ for the reduced loss \eqref{eq:F_A} and substituting $q=1-m^2$, the prefactor $(m^2+q)$ becomes $1$ and
\begin{align}
\label{eq:Phi_C}
    \Phi(m,b):=\mathcal{F}\!\left(m,1-m^2,b\right)=k(m,b)-2\,m\,h(m,b),
\end{align}
with $h,k$ the Gaussian averages \eqref{eq:hk_A} evaluated on \eqref{eq:U_C}. For the balanced case ($\omega=1/2$, $b=0$) relevant to each block of $(\mathcal D_2)$ these reduce to
\begin{align}
\label{eq:hk_C}
    h(m)=\mathbb{E}_g\!\left[\tanh\!\left(\gamma^2m+\gamma g\right)\right],
    \qquad
    k(m)=\mathbb{E}_g\!\left[\tanh^2\!\left(\gamma^2m+\gamma g\right)\right].
\end{align}
The problem is now one-dimensional: a single scalar $m$, and a single parameter $\gamma$.

\paragraph*{Constrained gradient flow.} Restricting the flow to the sphere $\lVert\vw\rVert=\alpha\sqrt d$ means projecting out the radial component, $\dd\vw/\dd\tau=-\left(\vI-\hat\vw\hat\vw^T\right)\nabla_{\vw}\mathcal{L}$ with $\hat\vw=\vw/\lVert\vw\rVert$. Using $\nabla_{\vw}m=\vmu/(\alpha d)$ and $\nabla_{\vw}q=2\vw^\perp/(\alpha^2d)$ from \eqref{eq:chain_A}, together with $\vw^T\nabla_{\vw}\mathcal{L}=m\,\partial_m\mathcal{L}+2q\,\partial_q\mathcal{L}$, one finds
\begin{align}
\label{eq:proj_flow_C}
    \frac{\dd m}{\dd\tau}
    =-\frac{1}{\gamma^2}\Big[\left(1-m^2\right)\partial_m\mathcal{L}-2mq\,\partial_q\mathcal{L}\Big]
    =-\frac{1-m^2}{\gamma^2}\Big[\partial_m\mathcal{L}-2m\,\partial_q\mathcal{L}\Big],
\end{align}
the second equality using $q=1-m^2$. The bracket is precisely the total derivative of the loss along the constraint surface, $\Phi'(m)=\partial_m\mathcal{F}-2m\,\partial_q\mathcal{F}$. Since $\mathcal{L}=\mathrm{const}+\alpha^2\beta^2\mathcal{F}+O_d(1/d^2)$ and $\alpha^2\beta^2/\gamma^2=\beta^2/d$, the flow closes on $m$ alone in the rescaled time $\check\tau=\beta^2\tau/d$ of \S\ref{app:unbalanced_exact},
\begin{align}
\label{eq:flow_C}
    \frac{\dd m}{\dd\check\tau}=-\left(1-m^2\right)\Phi'(m),
\end{align}
so that the dynamics again unfolds on $\tau=O_d(d)$. 

\paragraph*{Linearized dynamics.} Everything is controlled by the same two Gaussian integrals \eqref{eq:Theta_A} as in the unconstrained case, now evaluated at $q=1$ since the width in \eqref{eq:U_C} is fixed:
\begin{align}
\label{eq:theta_C}
    &\theta_1(\gamma):=\Theta_1(1,\gamma)=\mathbb{E}_g\!\left[\tanh^2\!\left(\gamma g\right)\right],\\
    &\theta_2(\gamma):=\Theta_2(1,\gamma)=\mathbb{E}_g\!\left[\left(1-\tanh^2\!\left(\gamma g\right)\right)\left(1-3\tanh^2\!\left(\gamma g\right)\right)\right].
\end{align}
Expanding \eqref{eq:hk_C} in $m$ around $0$ and using that $\tanh$ is odd (so $\mathbb{E}_g[\tanh(\gamma g)]=\mathbb{E}_g[\tanh\tanh'(\gamma g)]=0$),
\begin{align}
\label{eq:hk_expand_C}
    h(m)=\gamma^2\left(1-\theta_1\right)m+O(m^3),
    \qquad
    k(m)=\theta_1+\gamma^4\theta_2\,m^2+O(m^4),
\end{align}
so that $\Phi$ is even in $m$ and
\begin{align}
\label{eq:Phi_expand_C}
    \Phi(m)=\theta_1-\frac{\lambda(\gamma)}{2}\,m^2+O(m^4),
    \qquad
    \lambda(\gamma)=4\gamma^2\left(1-\theta_1(\gamma)\right)-2\gamma^4\,\theta_2(\gamma).
\end{align}
Inserting into \eqref{eq:flow_C}, the overlap grows exponentially from its $O_d(1/\sqrt d)$ initial value,
\begin{align}
\label{eq:rate_C}
    \frac{\dd m}{\dd\check\tau}=\lambda(\gamma)\,m+O(m^3).
\end{align}
One can check that $\lambda(\gamma)>0$ for all $\gamma>0$ and, unlike the unconstrained rate $\mu(q,\gamma)$ of \eqref{eq:lam_Xi_A}, which drifts as $q$ relaxes, $\lambda$ is a fixed number once $\gamma$ is chosen.

\paragraph*{Small $\gamma$ and consistency with \S\ref{app:hierarchical_exact}.} For $\gamma\ll1$, $\tanh(\gamma g)=\gamma g-\tfrac13\gamma^3g^3+O(\gamma^5)$ gives $\theta_1=\gamma^2-2\gamma^4+O(\gamma^6)$ and $\theta_2=1-4\gamma^2+O(\gamma^4)$, hence
\begin{align}
\label{eq:mu_small_C}
    \lambda(\gamma)=4\gamma^2\left(1-\gamma^2\right)-2\gamma^4+O(\gamma^6)=4\gamma^2-6\gamma^4+O(\gamma^6).
\end{align}
Applied blockwise to $(\mathcal D_2)$, where $\gamma_i^2=\kappa_i\Lambda\Gamma\to\kappa_i\Lambda$ by \eqref{eq:snr_B}, this gives
\begin{align}
\label{eq:mu_blocks_C}
    \lambda(\gamma_i)\simeq4\gamma_i^2=4\,\kappa_i\Lambda,
\end{align}
identical to the limiting rate \eqref{eq:lam_lim_B} obtained there without the constraint. 

Note that the two descriptions do not coincide away from this limit. 
For instance, a direct comparison at $q=1$ gives the exact identity
\begin{align}
\label{eq:compare_C}
    \lambda(\gamma)-\mu(1,\gamma)=2\left(\theta_1+\gamma^2\theta_2\right)=2\,\Psi(1,\gamma)>0.
\end{align}

\subsection{Integrated loss: SNR reparametrization and effective weightings}
\label{app:integrated_snr}

The ResNet experiments of Fig.~\ref{fig:gmm-resnet} do not compare two distinct pipelines. Both panels train the same denoiser, on the same denoising score matching objective \eqref{eq_DSM} and the same noising schedule; the two curves differ only through the weighting function $w(t)$, taken either uniform or chosen so as to reproduce the allocation of noise levels that Flow Matching performs implicitly. This appendix makes that construction precise. We first show that, once expressed in the signal-to-noise ratio, any stochastic interpolant pipeline is characterized by a single effective SNR weighting $w_{\mathrm{eff}}(\Lambda)$, so that two pipelines sharing it train identically whatever their time parametrization. We then compute $w_{\mathrm{eff}}$ in closed form for Diffusion and Flow Matching; the weighting $w(t)$ that installs the latter inside a DSM run is derived in \S\ref{app:fm_matching}, and is the one used in Fig.~\ref{fig:gmm-resnet}.

\paragraph*{Setup.} In practice a single network represents the denoiser at every noise level at once, and is trained on the loss integrated over the noising time with a weighting function $w(t)$,
\begin{align}
\label{eq:int_loss_D}
    \mathcal{L}(\vtheta)=\frac1d\int_0^{\tmax}\dd t\;w(t)\;\mathbb{E}_{\vx_0,\vxi}\left\lVert\vxi-\hat\vxi_{\vtheta}\left(\alpha(t)\vx_0+\beta(t)\vxi,\,t\right)\right\rVert^2 .
\end{align}
Here $\hat\vxi_{\vtheta}:=-\beta(t)\,\vs_{\vtheta}$ is the noise prediction associated with the score model, so that \eqref{eq:int_loss_D} is identically the loss $\lVert\beta\vs_{\vtheta}+\vxi\rVert^2$ of \eqref{eq_DSM} used in the single time analysis: the two are the same residual written in two parametrizations, for every $t$, with no approximation on $\beta$. The single SNR results of \S\ref{app:unbalanced_exact}--\S\ref{app:hierarchical_exact} may therefore be inserted into \eqref{eq:int_loss_D} as they stand.

\paragraph*{SNR reparametrization. } Two pipelines with different $(\alpha,\beta,w)$ may present the network with the same collection of denoising problems, since what a training example teaches depends on its noise level only through the SNR
\begin{align}
\label{eq:snr_D}
    \Lambda_t=\frac{\alpha^2(t)\,d}{\alpha^2(t)\sigma^2+\beta^2(t)},
\end{align}
which, assuming $\beta/\alpha$ monotonically increases, decreases monotonically from $\Lambda=d/\sigma^2$ at $t=0$ to $\Lambda=0$ at $t=\tmax$ (\S\ref{app:speciation}). Being monotone, $t\mapsto\Lambda_t$ is invertible and may be used as the integration variable in \eqref{eq:int_loss_D},
\begin{align}
\label{eq:int_loss_snr_D}
    \mathcal{L}(\vtheta)=\frac1d\int_0^{d/\sigma^2}\dd\Lambda\;w_{\mathrm{eff}}(\Lambda)\;\mathbb{E}_{\vx_0,\vxi}\left\lVert\vxi-\hat\vxi_{\vtheta}\right\rVert^2,
    \qquad
    w_{\mathrm{eff}}(\Lambda)=w\big(t(\Lambda)\big)\left\lvert\frac{\dd t}{\dd\Lambda}\right\rvert .
\end{align}
We call $w_{\mathrm{eff}}$ the effective SNR weighting of the scheme. Two pipelines with the same $w_{\mathrm{eff}}$ train identically whatever their time parametrizations; conversely, the same $w$ produces very different $w_{\mathrm{eff}}$ for different interpolants. All the comparisons below are therefore comparisons of $w_{\mathrm{eff}}$.

\paragraph*{Flow Matching as a reweighted denoising loss.} Diffusion regresses the noise directly, so \eqref{eq:int_loss_D} applies verbatim. Flow Matching instead regresses the velocity field along the linear interpolant $\vx_t=(1-t)\vx_0+t\vxi$, whose target is $\dot\vx_t=\vxi-\vx_0$,
\begin{align}
\label{eq:fm_loss_D}
    \mathcal{L}_{\mathrm{FM}}=\int_0^1\dd t\,w(t)\,\mathbb{E}\left\lVert\hat{\bm v}-\left(\vxi-\vx_0\right)\right\rVert^2 .
\end{align}
Eliminating $\vx_0=(\vx_t-t\vxi)/(1-t)$ gives $\vxi-\vx_0=(\vxi-\vx_t)/(1-t)$, so that with the change of variables $\hat\vxi=(1-t)\hat{\bm v}+\vx_t$,
\begin{align}
\label{eq:fm_to_dsm_D}
    \mathcal{L}_{\mathrm{FM}}=\int_0^1\dd t\;\frac{w(t)}{(1-t)^2}\;\mathbb{E}\left\lVert\hat\vxi-\vxi\right\rVert^2 .
\end{align}
Flow Matching is thus a denoising loss in disguise, but one that silently reweights the noise levels by $1/(1-t)^2=1/\alpha^2(t)$. This Jacobian is not a cosmetic detail: since $\alpha\to0$ at the noisy end, it is precisely what will concentrate FM on the low-SNR region below.

\paragraph*{Effective weightings of DM and FM.} We take $w\equiv1$; a non-uniform $w$ simply multiplies the results by $w(t(\Lambda))$.

For DM, $\alpha=e^{-t}$ and $\beta=\sqrt{1-e^{-2t}}$, so $\Gamma_t=1-e^{-2t}(1-\sigma^2)$ and \eqref{eq:snr_D} inverts as $e^{-2t}=\Lambda/(d+(1-\sigma^2)\Lambda)$. Differentiating $t=-\tfrac12\log e^{-2t}$,
\begin{align}
\label{eq:w_dm_D}
    w_{\mathrm{eff}}^{\mathrm{DM}}(\Lambda)=\frac{d}{2\Lambda\left(d+(1-\sigma^2)\Lambda\right)}
    \;\underset{\Lambda\ll d}{\simeq}\;\frac{1}{2\Lambda}.
\end{align}
For FM, $\alpha=1-t$ and $\beta=t$, so $\Gamma_t=(1-t)^2\sigma^2+t^2$ and, writing $r=(1-t)/t$, \eqref{eq:snr_D} reads $\Lambda=r^2d/(1+r^2\sigma^2)$, i.e.\ $r=\sqrt{\Lambda/(d-\sigma^2\Lambda)}$. Including the Jacobian of \eqref{eq:fm_to_dsm_D},
\begin{align}
\label{eq:w_fm_D}
    w_{\mathrm{eff}}^{\mathrm{FM}}(\Lambda)=\frac{d}{2\Lambda^{3/2}\sqrt{d-\sigma^2\Lambda}}
    \;\underset{\Lambda\ll d}{\simeq}\;\frac{\sqrt d}{2\Lambda^{3/2}}.
\end{align}
The two differ by a factor $\sqrt\Lambda$: $w_{\mathrm{eff}}^{\mathrm{FM}}/w_{\mathrm{eff}}^{\mathrm{DM}}\simeq\sqrt{d/\Lambda}$ for $\Lambda\ll d$, so FM up-weights low SNR ever more strongly as $\Lambda$ decreases.
Since the SNR spans several decades, it is enlightening to consider the alternative log-SNR coordinate $u=\log\Lambda$, $w_{\mathrm{eff}}\,\dd\Lambda=\tilde w(u)\,\dd u$ with
\begin{align}
\label{eq:logdens_D}
    \tilde w(u)=\Lambda\,w_{\mathrm{eff}}(\Lambda),
    \qquad
    \tilde w_{\mathrm{DM}}\simeq\frac12,
    \qquad
    \tilde w_{\mathrm{FM}}\simeq\frac12\sqrt{\frac d\Lambda},
\end{align}
for $\Lambda\ll d$. 
Under this change of variable, it clearly appears that DM is scale-free: it deposits the same weight on every decade of SNR, while FM is not: its weight per decade grows as $\sqrt{d/\Lambda}$ towards low SNR. Evaluated at speciation scale yields
\begin{align}
\label{eq:dens_ratio_D}
    \frac{\tilde w_{\mathrm{FM}}}{\tilde w_{\mathrm{DM}}}\bigg\rvert_{\Lambda=O_d(1)}
    =O_d\!\left(\sqrt d\right).
\end{align}

\subsection{Matching Diffusion and Flow Matching at equal SNR}
\label{app:fm_matching}

The comparison of \S\ref{app:integrated_snr} is between weightings, not between implementations. To isolate that effect experimentally we keep everything else fixed -- same $\vxi$ prediction network, same VP forward process $\vx_t=e^{-t}\vx_0+\sqrt{1-e^{-2t}}\,\vxi$, same sampler -- and change only the weighting $w(t)$ in \eqref{eq:int_loss_D}: the baseline uses $w\equiv1$, i.e.\ plain DSM, and the ``FM-weighted'' run uses the $w(t)$ that reproduces the effective SNR weighting of Flow Matching. We derive that weight here.

\paragraph*{The matched time.} Write $u(t)=\sqrt{e^{2t}-1}$, so that the VP interpolant has $\alpha_{\mathrm{DM}}/\beta_{\mathrm{DM}}=e^{-t}/\sqrt{1-e^{-2t}}=1/u$. The linear interpolant $\vx_r=r\vx_0+(1-r)\vxi$ has $\alpha_{\mathrm{FM}}/\beta_{\mathrm{FM}}=r/(1-r)$. Since $\Lambda$ is a monotone function of $\alpha/\beta$ alone, the two processes present the same denoising problem when
\begin{align}
\label{eq:tau_match}
    \frac{r}{1-r}=\frac1u
    \qquad\Longleftrightarrow\qquad
    r(t)=\frac{1}{1+u(t)} .
\end{align}

Differentiating \eqref{eq:tau_match} with $\dd u/\dd t=e^{2t}/u$ gives $\lvert\dd r/\dd t\rvert=e^{2t}/\big(u(1+u)^2\big)$. The two combine into
\begin{align}
\label{eq:w_match}
    w_{\mathrm{FM}\to\mathrm{DM}}(t)
    =\left(1+u\right)^2\times\frac{e^{2t}}{u\left(1+u\right)^2}
    =\frac{e^{2t}}{\sqrt{e^{2t}-1}}.
\end{align}
One checks directly that uniform $t$ sampling with this weight induces exactly $w_{\mathrm{eff}}^{\mathrm{FM}}$ of \eqref{eq:w_fm_D}. The two limits are the ones anticipated in \S\ref{app:integrated_snr}: at large $t$, $u\simeq e^{t}$ and $w_{\mathrm{FM}\to\mathrm{DM}}\simeq e^{t}=\sqrt{d/\Lambda}$, recovering the tilt of \eqref{eq:logdens_D}; at small $t$ it diverges as $1/\sqrt{2t}$, which is regularized in practice by the cutoff $t_{\min}>0$ used in training.

\subsection{Masked discrete diffusion}
\label{app:mdlm_snr}

The SNR construction of \S\ref{app:integrated_snr} is not tied to Gaussian interpolants. We treat here the masked discrete diffusion (MDLM) pipeline of \citet{sahoo2024mdlm} used on the genomic data of \S\ref{app:mdlm-humgen}.

\paragraph{Noise injection and SNR.} In this case, each of the $d$ tokens of $\vx_0$ is independently replaced by \texttt{MASK} with probability $1-e^{-\sigma(t)}$, so that a token survives with probability $e^{-\sigma(t)}$, and $\sigma$ increases from $\sigma_{\min}$ at $t=0$ to $\sigma_{\max}$ at $t=1$.

In the discrete masking paradigm, a masked token carries no information at all, while an unmasked one carries it intact: masking destroys information by removing tokens rather than by attenuating them. The natural signal-to-noise ratio is therefore simply the expected number of tokens that survive,
\begin{align}
\label{eq:snr_mdlm_D}
    \Lambda_t=e^{-\sigma(t)}\,d ,
\end{align}
which plays exactly the role that $\Lambda_t=\alpha^2d/\Gamma_t$ plays in the Gaussian case,
both counting effective informative units (we absorb the $O_d(1)$ information per token into the normalization). 

\paragraph{Effective weighting.} As implemented, the ELBO \eqref{eq:app-elbo} averages the cross-entropy over the masked positions and reweights it by $\dot\sigma/(1-e^{-\sigma})$, so that the per-token loss carries the time weighting $w(t)=\dot\sigma/(1-e^{-\sigma})$. With $\Lambda=e^{-\sigma}d$ we have $\lvert\dd t/\dd\Lambda\rvert=1/(\dot\sigma e^{-\sigma}d)$, and the $\dot\sigma$ cancels:
\begin{align}
\label{eq:w_mdlm_D}
    w_{\mathrm{eff}}^{\mathrm{MDLM}}(\Lambda)=\frac{\dot\sigma}{1-e^{-\sigma}}\cdot\frac{1}{\dot\sigma\,e^{-\sigma}d}
    =\frac{d}{\Lambda\left(d-\Lambda\right)}
    \;\underset{\Lambda\ll d}{\simeq}\;\frac1\Lambda .
\end{align}
Two consequences follow. First, \eqref{eq:w_mdlm_D} does not involve $\sigma(t)$: the effective weighting is independent of the noise schedule, a discrete counterpart of the known schedule invariance of the masked diffusion ELBO. Only the endpoints $\sigma_{\min},\sigma_{\max}$ matter, through the range of $\Lambda$ they expose. Second, comparing with \eqref{eq:w_dm_D}, MDLM has the same $1/\Lambda$ low SNR behavior as Diffusion: measured in SNR, masked discrete diffusion is scale-free and  rather trains like a standard DSM than a FM.

\section{Numerical details}
\label{app:numerical_details}

\subsection{Fig. \ref{fig:exscore-unbalanced} and Fig. \ref{fig:exscore-kappa}}

In these experiments, we consider synthetic Gaussian-mixture data in
$\mathbb{R}^d$ as described in the main text, focusing on two datasets:
\begin{itemize}
  \item \textbf{Unbalanced} (two-mode GMM): a single signal direction
  $\vmu$ with unequal mode probability $\omega=0.8$, swept over the
dimension $d\in\{64,128,256\}$ at fixed $\omega$.
  \item \textbf{Kappa dependence} (Quadrimodal GMM): two orthogonal
  signal directions $\vmu_1\perp\vmu_2$ of unequal strength ($\kappa$ vs.\
  $1-\kappa$), each with its own mode probability $\omega_1=\omega_2=0.7$
  (asymmetric $\pm$ sign per mode). $d=1024$ fixed, swept over
  $\kappa\in\{0.1,0.2,0.3\}$.
\end{itemize}
In these experiments the score's functional form is known exactly from the Gaussian-mixture log-density at all noise levels, and only a handful of scalar/vector
coefficients in that closed form are learned, by denoising score
matching (DSM) of a VP-SDE at a single, fixed noise level. In both settings we compare two fixed DSM
noise times $t\in\{0.5,\,t_s\}$ side by side, where $t_s=\tfrac12\log d$. 
There is no
sampling or sample generation anywhere in either experiment: because the
score's closed form is known exactly, model performance is evaluated
directly from the learned coefficients themselves, rather than from
samples generated by the model.

\subsubsection{Data}
\label{sec:data-exact}
During training, data samples $\vx$ are generated as follows:

\paragraph*{Unbalanced GMM (dimension sweep).} $\vx=s\vmu+\vz$, $\vmu=\mathbf{1}_d$
(all-ones, so $\lVert\vmu\rVert^2=d$), $\vz\sim\mathcal{N}(0,\vI_d)$, $s=+1$
with probability $\omega=0.8$ (else $-1$). Swept over
$d\in\{64,128,256\}$ at fixed $\omega=0.8$.

\paragraph*{Quadrimodal GMM (kappa-dependence sweep).}
$\vx = s_1\vmu_1 + s_2\vmu_2 + \vz$, $\vz\sim\mathcal{N}(0,\vI_d)$, $d=1024$.
$\vmu_1,\vmu_2$ are orthogonal block-indicator vectors
($\lVert\vmu_1\rVert^2=k_1=\lfloor\kappa d\rfloor$,
$\lVert\vmu_2\rVert^2=k_2=\lfloor(1-\kappa)d\rfloor$), with unbalanced
signs: $s_1=+1$ with probability $\omega_1=0.7$ (else $-1$), independently
$s_2=+1$ with probability $\omega_2=0.7$. Swept over $\kappa\in\{0.1,0.2,0.3\}$.

\subsubsection{Model architecture}
\label{sec:model-exact}

As discussed in depth in the main text, both experiments exploit closed-form knowledge of the true score,
parameterizing it with a small number of scalar/vector coefficients
rather than a generic function approximator; metrics are read directly
off these learned coefficients.

\paragraph*{Unbalanced.} The exact score of the two-mode mixture factorizes
as
\begin{equation}
\nabla_{\vx}\log P_t(\vx) = -c\,\vx + \vw\tanh(\vx^T \vw + b),    
\end{equation}

with $\vw\in\mathbb{R}^d$ playing the role of the learned $\vmu$-direction, $c$ the pseudo variance, and the log weight $b\to\tfrac12\log\frac{\omega}{1-\omega}$ at convergence. The model thus has $(d+2)$ learnable parameters in total.

\paragraph*{Quadrimodal.} Because the four modes have equal weight (given
$s_1,s_2$) and $\vmu_1\perp\vmu_2$ by construction, the exact score of the
VP-SDE-noised mixture at a fixed time $t$ similarly factorizes:
\[
\nabla_{\vx}\log P_t(\vx) = -c\,\vx + \vw_1\tanh(\vx^T \vw_1+b_1) + \vw_2\tanh(\vx^T \vw_2+b_2),
\]
with $\vw_1\in\mathbb{R}^{k_1}$, $\vw_2\in\mathbb{R}^{k_2}$ each compactly
supported on their own mode's subspace, plus scalars $c,b_1,b_2$:
$d{+}3$ learnable parameters total. 

Both models use a small initialization: $\vw_i \sim d^{-\frac{1}{2}}\mathcal{N}(0, \vI_{d})$ (respectively $\vw_i \sim k_i^{-\frac{1}{2}}\mathcal{N}(0, \vI_{k_i})$), such that initially $\lVert\vw_i\rVert^2 = O(1)$

\subsubsection{Training procedure}
\label{sec:training-exact}

For each fixed DSM time $t$, we let $\alpha_t=e^{-t}$, $\sigma_t=\sqrt{1-e^{-2t}}$, and optimize the loss 

\begin{equation}
    \mathcal{L}=\mathbb{E}\lVert\sigma_t\,{\rm
score}(\vx_t)+\vxi\rVert^2
\end{equation}
where $\vx_t=\alpha_t \vx_0 + \sigma_t\vxi$ and the expectation is taken over training data $\vx_0$ and $\vxi \sim \mathcal{N}(0, \vI_d)$.

\subsubsection{Unbalanced, dimension dependence, Fig.~\ref{fig:exscore-unbalanced}}
\label{sec:fig-unbalanced-exact}

The two panels are the two fixed noising times: $t=0.5$ \textit{(left)}, which sits deep in \textbf{Regime~II}, and $t=t_s=\tfrac12\log d$ \textit{(right)}, the \textbf{speciation} time. colors index $d\in\{64,128,256\}$, and the three curves of the legend are, for each $d$:
\begin{itemize}
  \item \textbf{Overlap} - the direction overlap $\mathbb{E}_+[\vw^T\vmu]/(e^{-t}d)$,
  normalized to $1$ at convergence.
  \item \textbf{c} - the learned scalar $c$, which plays the
  role of an inverse-variance-like precision term in the exact-score
  formula.
  \item \textbf{weight} - the learned mode weight $\omega = 0.5+\lvert\operatorname{sigmoid}(2b)-0.5\rvert$, read off the bias. The dotted horizontal line marks its target value $\omega=0.8$.
\end{itemize}
The $x$ axis is the number of SGD steps divided by $d$.

\subsubsection{Kappa dependence, Fig.~\ref{fig:exscore-kappa}}
\label{sec:fig-kappa-exact}

The figure has three panels: $t=0.5$ \textit{(left)}, deep in \textbf{Regime~II}; $t=t_s$ \textit{(middle)}, at \textbf{speciation}; and the same speciation data replotted against a rescaled $x$ axis \textit{(right)}. colors index $\kappa\in\{0.1,0.2,0.3\}$, and for each $\kappa$ the two curves of the legend are the overlaps along the two orthogonal directions,
\begin{align}
  m_1=\frac{\mathbb{E}_+[\vw_1^T\vmu_1]}{e^{-t}\kappa d},
  \qquad
  m_2=\frac{\mathbb{E}_+[\vw_2^T\vmu_2]}{e^{-t}(1-\kappa)d},
\end{align}
each normalized to $1$ at convergence ($m_1$ dashed, $m_2$ solid). The learned biases $b_i$ are not displayed here.

The $x$ axis of the first two panels is the raw number of SGD steps. In the third it is rescaled by the growth rate of the corresponding block, written $\lambda(\sqrt{\kappa_i})$ in the figure and equal to the closed-form rate $\lambda(\gamma_i)$ of~\eqref{eq:mu_main}, evaluated at $\gamma_i^2=\kappa_i\Lambda_t\Gamma_t$ with $\kappa_1=\kappa$ and $\kappa_2=1-\kappa$.

\paragraph*{Relation to $(\mathcal{D}_2)$.} The hierarchical target
\eqref{eq:hierarchical} of the main text is balanced, all four modes carrying
weight $1/4$, whereas the runs here draw each sign with probability
$\omega_1=\omega_2=0.7$. The exact score model keeps one bias per block, $b_1,b_2$,
initialized at zero, so this imbalance is representable and is learned
alongside the overlaps. Strictly, each block then follows the coupled
$(m_i,b_i)$ dynamics of Result~\ref{thm:unbalanced} rather than the balanced
reduction of Result~\ref{thm:hierarchical}, the off-diagonal coupling being
proportional to $2\omega_i-1$. This does not affect what the figure tests: the
growth rate of block $i$ remains a function of $(q_i,\gamma_i)$ alone, hence
of $\kappa_i\Lambda$, so the collapse variable is unchanged and the balanced
closed form $\lambda(\gamma_i)$ is used as the rescaling ansatz. The ResNet
experiments below instead use the balanced target exactly.

Training hyperparameters are reported below:

\begin{table}[h!]
\centering
\begin{tabular}{lll}
\toprule
 & \multicolumn{2}{c}{Unbalanced and Kappa-dependence} \\
\midrule
Optimizer & \multicolumn{2}{c}{plain SGD (no momentum, no weight decay, constant LR)} \\
Learning rate & \multicolumn{2}{c}{$10^{-1}$} \\
Batch size & \multicolumn{2}{c}{$512$} \\
Training steps & \multicolumn{2}{c}{$10^{6}$ (SGD steps, no early stopping)} \\
$t_{\rm noise}$ & \multicolumn{2}{c}{$0.5$ \quad or \quad $t_s=\tfrac12\log d$} \\
Seeds per config & \multicolumn{2}{c}{$20$}  \\
Sweep values & $d\in\{64,128,256\}$ at $\omega=0.8$ & $\kappa\in\{0.1,0.2,0.3\}$ at $d=1024$ \\
\bottomrule
\end{tabular}
\caption{Hyperparameters shared by all $20$ runs averaged to produce each figure. Metrics are logged at $200$ log-spaced steps.}
\end{table}

\subsection{Fig. \ref{fig:gmm-resnet-weight-learning} and Fig. \ref{fig:gmm-resnet-struct-learning}}
\label{sec:dsm-sibling}

In these experiments, we return to the same two Gaussian-mixture datasets
introduced above — unbalanced and quadrimodal GMMs, with data generated
as in \S\ref{sec:data-exact} except that the quadrimodal signs are now drawn
with equal probability, $\omega_1=\omega_2=0.5$, so that the target is exactly the
balanced $(\mathcal{D}_2)$ of \eqref{eq:hierarchical}; the unbalanced target
keeps $\omega=0.8$ — but replace the analytically exact
score parameterization with a generic fully-connected ResNet
epsilon-predictor, trained by denoising score matching (DSM) over a
continuous range of noise levels rather than at a single fixed $t$.
Correspondingly, since the score is no longer known in closed form,
model performance can no longer be read directly off a handful of
coefficients: instead, we evaluate it by drawing samples from the model
via reverse-time SDE integration and comparing statistics of those
samples to the training distribution. 
We contrast two training objectives, standard DSM and FM-weighted DSM.

\subsubsection{Model architecture}
\label{sec:model-gmm}

Both experiments use the same fully-connected ResNet epsilon-predictor
$\hat\vxi_{\vtheta}(\vx,t): \mathbb{R}^d\times\mathbb{R}\to\mathbb{R}^d$: a
linear input projection to width $128$, $4$ residual blocks (LayerNorm,
additive sinusoidal-time-embedding conditioning via a linear projection,
then a $2$-layer SiLU MLP, residual add), and a zero-initialized linear
output projection back to $\mathbb{R}^d$. Sinusoidal time embedding
dimension $128$. The score is reparameterized with a skip
connection. The input/output projections are the only $d$-dependent
layers.

\subsubsection{Training procedure}
\label{sec:training-gmm}

In both experiments we follow the standard denoising score matching
procedure, where noised samples are obtained via the forward
continuous-time variance-preserving Ornstein-Uhlenbeck SDE: $\text{d}\vx_t=-\vx_t\,\text{d}t+
\sqrt{2}\,{\text d}\vb_t$, with closed-form marginal $\vx_t\mid \vx_0 \sim
\mathcal{N}(e^{-t}\vx_0, \sqrt{(1-e^{-2t})}\vI)$. The network $\hat\vxi_{\vtheta}$
predicts the injected noise $\vxi$, and the score used in the
backward inference dynamics is recovered via Tweedie's formula,
$\nabla_{\vx}\log p_t(\vx) = -\hat\vxi_{\vtheta}(\vx,t)/\sigma(t)$. Sampling
integrates the reverse SDE with Euler–Maruyama from
$t_{\max}$ down to $t_{\min}>0$.

To evaluate the impact of weighting schemes, we use a weighted denoising
score matching loss
\begin{equation}
    \label{eq:app-weight-dsml}
    \mathcal{L}=\mathbb{E}_{t\sim\mathcal{U}(t_{\min},t_{\max})}\big[w(t)\,
\lVert\vxi-\hat\vxi_{\vtheta}(\vx_t,t)\rVert^2\big]
\end{equation}
with two choices of $w(t)$:
\begin{itemize}
  \item \textbf{Uniform DSM}: $w(t)=1$.
  \item \textbf{FM-weighted DSM}: $w(t)=e^{2t}/\sqrt{e^{2t}-1}$, the matched
  weight \eqref{eq:w_match} derived in \S\ref{app:fm_matching}. 
\end{itemize}
For each training we use plain SGD with a constant learning rate and
neither momentum nor weight decay. 
The curves report the mean over the generated samples at each logged
step and the shaded band the corresponding $95\%$ CI, so the band measures
sampling error at fixed model rather than run-to-run variability. 
Inference and training hyperparameter values are reported below:

\begin{table}[h!]
\centering
\begin{tabular}{lll}
\toprule
 & Unbalanced GMM & Quadrimodal GMM \\
\midrule
$d$ & $64$ & $64$ \\
Optimizer & \multicolumn{2}{c}{plain SGD (no momentum, no weight decay, constant LR)} \\
Learning rate & $10^{-2}$ & $10^{-2}$ \\
Batch size & $2048$ & $2048$ \\
Training steps (SGD) & $10^6$ & $10^6$ \\
$t_{\min}, t_{\max}$ & $5\times10^{-3},\ 6.2$ & $10^{-3},\ 6.2$ \\
Reverse SDE steps (eval) & $30$ & $30$ \\
Eval samples / logged step & $5\times10^4$ &  $5\times10^4$ \\
Sweep values & $\omega=0.8$ & $\kappa\in\{0.2,0.3,0.4\}$ \\
\bottomrule
\end{tabular}
\caption{Hyperparameters associated to each figure. Both objectives
(uniform / FM-weighted DSM) use identical hyperparameters otherwise.}
\end{table}

\subsubsection{Unbalanced GMM, Fig. \ref{fig:gmm-resnet-weight-learning}}
\label{sec:fig-weight-gmm}

The two panels are the two weighting schemes of \eqref{eq:app-weight-dsml}, labeled \textbf{Uniform DSM} ($w(t)=1$) and \textbf{FM-weighted DSM}, at the single imbalance $\omega=0.8$ marked by the dotted horizontal line. For each we monitor the two curves of the legend:
\begin{itemize}
    \item \textbf{Cosine Similarity} - the cosine between an inferred sample and the mode direction, $\cos(\vx,\vmu)=\vx^T\vmu/(\lvert \vx\rvert\lvert\vmu\rvert)$. Since the target is symmetric, we average the mean cosine over the two signs of $\vx^T\vmu$ and report $\tfrac12\big(\lvert\mathbb{E}_+[\cos(\vx,\vmu)]\rvert+\lvert\mathbb{E}_-[\cos(\vx,\vmu)]\rvert\big)$. This tracks direction learning: a transition to $O(1)$ values indicates that the inferred samples are spread along the preferential training direction $\vmu$. Note that this does not mean the inferred distribution is necessarily bimodal, only that its covariance matrix is no longer isotropic.
    \item \textbf{Positive fraction} - the fraction of generated samples falling on the dominant side of the mode direction, $\tfrac1N\#\{\vx:\vx^T\vmu>0\}$. It converges to $\omega$ once the imbalance has been acquired.
\end{itemize}
Each panel carries an inset showing the histogram of the projection $p=\vx^T\vmu/\lVert\vmu\rVert^2$ of the generated samples, at the three training times marked by the vertical dashed lines; the dotted verticals mark the target mode locations $\pm 1$.

\subsubsection{Quadrimodal GMM, Fig. \ref{fig:gmm-resnet-struct-learning}}
\label{sec:fig-struct-gmm}

Again the two panels are the two weighting schemes, \textbf{Uniform DSM} and \textbf{FM-weighted DSM}, and colors index $\kappa\in\{0.2,0.3,0.4\}$. The monitored quantity is the cosine similarity between the inferred samples and the subdominant direction $\vmu_1$, defined as in \S\ref{sec:fig-weight-gmm} with $\vmu$ replaced by $\vmu_1$; since $\lVert\vmu_1\rVert^2=\kappa d$, smaller $\kappa$ means a weaker subdominant mode.

Only the \textbf{FM-weighted DSM} panel carries an inset, which replots that panel's curves against the rescaled $x$ axis, training steps $\times\,\lambda(\sqrt{\kappa})$, with $\lambda$ the closed-form rate \eqref{eq:mu_main} evaluated at $\gamma_i^2=\kappa_i\Lambda$.

\subsection{Fig. \ref{fig:mnist-weight-learning} and Fig. \ref{fig:mnist-hierarchical}}

We extend the ResNet/GMM experiments with two companion experiments on MNIST digits $\{3,6\}$, sharing
the same generative architecture and varying a different
axis of the data-generating process:
\begin{itemize}
  
  \item \textbf{Weight learning} (class-imbalanced grayscale MNIST): only
  digit identity varies, but the two classes appear in an imbalanced
  ratio $\omega = P(\text{class}=6)$ in the training set. 

  \item \textbf{Structure learning} (tinted MNIST): both digits are shown
  in a random, class-independent color (red or green) at a fixed contrast
  $\alpha$. The data has two orthogonal directions of variation — digit
  identity (``3 vs.\ 6'') and color (``red vs.\ green'') — and $\alpha$
  controls the strength of the (weaker) color signal. 
\end{itemize}

\subsubsection{Data}
\label{sec:data-mnist}

Both experiments restrict MNIST (train split) to digits 3 and 6
($n_3=6{,}131$, $n_6=5{,}918$ images, $28\times28$).

\paragraph*{Class-imbalanced MNIST.} Grayscale (1 channel),
normalized to $[-1,1]$. All $n_6$ sixes are kept and threes are
subsampled to hit a target mixture weight $\omega\in\{0.5,0.8\}$:
$n_{\mathrm{tot}} = \lfloor n_6/\omega\rfloor$, $n_3 =
n_{\mathrm{tot}}-n_6$ (so $\omega=0.5$: balanced, $n_3=n_6=5{,}918$;
$\omega=0.8$: $n_3=1{,}479$, $n_6=5{,}918$, total $7{,}397$).

\begin{figure}[h!]
    \centering
    \includegraphics[width=0.5\figwidth]{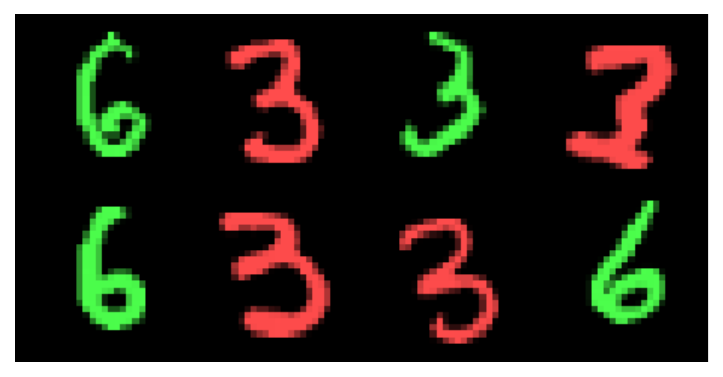}
    \caption{Image sample at tint $\alpha = 0.7$.}
    \label{fig:app-tint-ex}
\end{figure}

\paragraph*{Tinted MNIST.} All $12{,}049$ available 3s and
6s are used, no subsampling. Each grayscale image $\vx\in[0,1]^{28\times
28}$ is expanded to 3 channels and randomly tinted red or green
(uniformly, independently of digit label) at contrast
$\alpha\in\{0.4,0.5,0.6,0.7,0.8,0.9,1.0\}$:
\begin{equation}
  \vx_{\mathrm{tinted}} = (1-\alpha)\,\bar{\vx} + \alpha\,\bar{\vx}\odot c,
  \qquad \bar{\vx} = (\vx,\vx,\vx),\quad c\in\{(1,0,0),(0,1,0)\},
\end{equation}
then normalized to $[-1,1]$ per channel. $\alpha=0$ would give plain
grayscale (no hue signal); $\alpha=1$ gives a fully saturated tint; see Fig. \ref{fig:app-tint-ex} for a sample.\\

\subsubsection{Model architecture}
\label{sec:model-mnist}

Both experiments use the same U-Net backbone
(\texttt{torchcfm.models.unet.UNetModel}): model channels $32$, $2$ residual
blocks per resolution for the imbalance experiment and $1$ for the tinted one,
channel multiplier $(1,2,2)$ (the package's default for $28\times28$
inputs), self-attention at the coarsest resolution (single head), no
class conditioning, dropout $0$, GroupNorm+SiLU residual blocks with a
sinusoidal timestep embedding. Input/output channels are $3$ for the
tinted experiment and $1$ for the grayscale one, both $\approx
1.08\times10^{6}$ parameters.

That backbone is used in two different roles. Under Flow Matching it is the
velocity field $\vv_{\vtheta}(\vx,t)$ regressed in \eqref{eq:app-CFM}. Under
DSM it parametrizes the score through a trainable skip connection,
\begin{align}
\label{eq:app-score-skip}
    \vs_{\vtheta}(\vx,t)=-\vx+\mathrm{UNet}_{\vtheta}(\vx,t),
\end{align}
rather than the usual $-\hat\vxi_{\vtheta}(\vx,t)/\sigma(t)$: at initialization
the network output is small, so the score defaults to $-\vx$, the score of a
standard normal, instead of an untrained noise prediction divided by a
possibly tiny $\sigma(t)$. This keeps reverse-SDE sampling stable before the
network has learned anything. 

\subsubsection{Flow-matching training procedure}
\label{sec:training-mnist}

In the colorized experiment, we use the standard conditional Flow Matching (CFM): for a clean sample $\vx_1$ and $\vx_0\sim\mathcal{N}(0,\vI)$,
\begin{equation}
  \vx_t = (1-t)\vx_0 + t \vx_1, \qquad t\sim\mathcal{U}(0,1),
\end{equation}
and the network is trained to regress the constant conditional velocity
$\dot \vx_t = \vx_1-\vx_0$ with an MSE loss,

\begin{equation}
\label{eq:app-CFM}
\mathcal{L}(\vtheta)=\mathbb{E}_{t,\vx_0,\vx_1}\lVert \vv_{\vtheta}(\vx_t,t)-(\vx_1-\vx_0)\rVert^2
\end{equation}
Sampling integrates $\vv_{\vtheta}$ from $t=0$ (Gaussian noise) to $t=1$ with
an adaptive-step Dormand–Prince (\texttt{dopri5}) ODE solver
(\texttt{torchdyn}, $\mathrm{atol}=\mathrm{rtol}=10^{-4}$). The training and inference parameters are reported below:

\begin{table}[h!]
\centering
\begin{tabular}{lll}
\toprule
 & Structure sweep (tinted) & Weight sweep (imbalanced) \\
\midrule
Optimizer & \multicolumn{2}{c}{AdamW, weight decay $10^{-2}$ (default), constant LR} \\
Learning rate & \multicolumn{2}{c}{$10^{-3}$} \\
Batch size & \multicolumn{2}{c}{$128$} \\
Training steps & $2{,}000$ & $10{,}000$ \\
Sweep values & $\alpha\in\{0.4,\dots,1.0\}$ (step $0.1$) & $\omega=0.8$ \\
Seeds per value & $10$ & $5$\\
Eval samples / logged step & $10{,}000$ ($1{,}000\times10$ repeats) & $5{,}000$ ($1{,}000\times5$ repeats) \\
Classifier epochs & $6$ & $20$ \\
\bottomrule
\end{tabular}
\caption{Hyperparameters for the two experiments.}
\label{table:mnist}
\end{table}

The imbalance experiment is run twice, once with the Flow Matching objective
above and once with the score-based DSM objective of \S\ref{sec:training-gmm},
which is the pair compared in Fig.~\ref{fig:mnist-weight-learning}. 
The DSM side uses the same U-Net with the score skip connection \eqref{eq:app-score-skip}, a unit-rate Ornstein-Uhlenbeck forward process
$\dd\vx=-\vx\,\dd t+\sqrt2\,\dd \vb_t$ run over $t\in[5\times10^{-3},3.3]$, and
generates by integrating the reverse SDE over $50$ steps with cutoff
$\epsilon_t=5\times10^{-3}$. All optimization
hyperparameters are those of Tab.~\ref{table:mnist}.

\subsubsection{Weight learning, Fig.~\ref{fig:mnist-weight-learning}}
\label{sec:fig-weight-mnist}

Both panels show the same unbalanced dataset at fixed $\omega=0.8$ (dotted
horizontal line); what differs is the training objective, \textbf{DSM} on the
left and \textbf{Flow Matching} on the right, so the two panels isolate the
effect of SI pipeline at fixed data. Curves are the mean over the $5$ seeds,
the band a $95\%$ CI across seeds, Gaussian-smoothed along the step axis.

At each logged step we draw samples from the current model and pass them
through two fixed evaluation probes, neither of which is ever used to train
the generative model:
\begin{itemize}
  \item \textbf{An auxiliary digit classifier}, a 1-channel, 2-layer CNN
  trained once per run on real data with a matching noise-augmentation scheme
  (classify $(1-t)\vx_0+t\vx_1$ against the true label,
  $t\sim\mathcal{U}(0,1)$, so it stays calibrated at the high-noise end of the
  path). It is always fit on a balanced ($\omega=0.5$) split, so it judges
  generated class frequency without inheriting the training-set imbalance.
  \item \textbf{Class-mean projections.} $\vmu_3,\vmu_6$, the pixel-space mean
  image of each digit class in the training set, precomputed per run; every
  generated sample $\vx$ is projected onto both, $\vx^T\vmu_3$ and
  $\vx^T\vmu_6$.
\end{itemize}
The two curves of the legend are then:
\begin{itemize}
    \item \textbf{Cosine Similarity} - $\mathbb{E}\big[\max\big(\cos(\vx,\vmu_3),\cos(\vx,\vmu_6)\big)\big]$, the cosine between a generated sample and the nearer of the two class-mean images. 
    \item \textbf{Fraction of 6s} - $\mathbb{E}[\mathbf{1}_{C(\vx)=6}]$, the empirical fraction of generated samples the classifier assigns to digit $6$. 
\end{itemize}

\subsubsection{Structure learning, Fig.~\ref{fig:mnist-hierarchical}}
\label{sec:fig-struct-mnist}

Two panels, the mean over the $10$ seeds per $\alpha$, the band a $95\%$ CI
across seeds, Gaussian-smoothed along the step axis. The digit direction is
read off the same class-mean projections as above; the color direction uses a
hue projection, the mean red minus mean green pixel intensity of a generated
image, $r-g$ (``color spread'').
\begin{itemize}
    \item Left panel - the digit direction overlap, defined as  $\mathbb{E}\big[\max(\frac{|\vx^T\vmu_3|}{|\vmu_3|^2},\frac{|\vx^T\vmu_6|}{|\vmu_6|^2})\big]$.
    \item Right panel - the color spread. Since $r-g$ is bimodal by construction we keep only its positive part and monitor $\mathbb{E}_+[\alpha\!\cdot\!(r-g)]/\alpha^2$. The $\alpha^{-2}$ normalization reflects the squared norm of the training color spread scaling as $\alpha^2$, mirroring the $|\vmu_{3,6}|^2$ normalization of the digit direction.
\end{itemize}
To rescale the time axis of the subdominant direction we identify the relative
direction amplitude $|\vmu_1|^2/|\vmu_2|^2$ as proportional to $\alpha^2$;
rescaling training time by \eqref{eq:mu_main} then collapses the curves.

\subsection{Fig. \ref{fig:mdlm-humgen}}
\label{app:mdlm-humgen}

\subsubsection{Dataset}

We use the 805-SNP subset of phased haplotypes from the 1000 Genomes
Project (\href{https://gitlab.inria.fr/ml_genetics/public/artificial_genomes/-/tree/master/1000G_real_genomes}{805\_SNP\_1000G\_real.hapt}). Each of the $2{,}504$
sequenced individuals contributes two phased haplotypes, giving
$N=5{,}008$ binary sequences of length $L=805$,
$\vx \in \{0,1\}^{L}$.
The data is split into a $90/10$ training and validation set.
A reference PCA basis $\{\vmu_k\}_{k=1}^{K}$,
$K=50$, together with the reference eigenvalues
$\{\lambda_k\}_{k=1}^{K}$, is fit once on the training split and kept
fixed throughout training for evaluation purposes.

\subsubsection{Masked Discrete Diffusion process}
\label{sec:diffusion-genome}

We use continuous-time absorbing-state (masked) discrete diffusion (MDLM)
\citep{sahoo2024mdlm}, adapted to a binary alphabet. Each token from the original sequence $\vx_0$ is
independently replaced in the noised sequence $\vx_t$ by \textbf{MASK} with probability
$\alpha(t) = 1-e^{-\sigma(t)}$, under a log-linear noise schedule
\begin{equation}
  \sigma(t) = \sigma_{\min}^{1-t}\,\sigma_{\max}^{\,t},
  \qquad t\in[0,1],
\end{equation}
with $\sigma_{\min}=10^{-4}$, $\sigma_{\max}=20$. The model is trained
with the continuous-time ELBO (masked cross-entropy at masked
positions only, reweighted by $\dot\sigma(t)/(1-e^{-\sigma(t)})$),
with $t \sim \mathcal{U}(\vxi, 1)$, $\epsilon=10^{-4}$, resampled
independently for every sequence in a minibatch: 

\begin{equation}
\label{eq:app-elbo}
    \mathcal{L}(\vtheta) = -\int_{\epsilon}^1 \frac{\dot\sigma(t)}{(1-e^{-\sigma(t)})}\mathbb{E}_{\vx_t, \vx_0}\left[\log(p_{\vtheta}(\vx_0|\vx_t))\right].
\end{equation}
Sampling uses the ancestral reverse process as originally proposed in the MDLM paper, starting from a fully masked
sequence and unmasking tokens over $n_{\mathrm{steps}}=30$ discretized
reverse steps. At each logged step, $10 \times 500$ samples are drawn from the current model to evaluate PCA overlap statistics.

\subsubsection{Model architecture}
\label{sec:model-genome}

The denoiser $p_{\vtheta}$ is a small Transformer encoder,
operating on a $3$-token vocabulary
$\{0,1,\mathrm{MASK}\}$:
\begin{itemize}
  \item \textbf{Input embedding.} A learned token embedding
  ($3 \times d_{\mathrm{model}}$) plus a learned absolute positional
  embedding ($L \times d_{\mathrm{model}}$), summed.
  \item \textbf{Time conditioning.} The noise level enters through
  $\log\sigma(t)$, mapped by a sinusoidal embedding of dimension
  $d_{\mathrm{cond}}$ followed by a $2$-layer MLP
  ($d_{\mathrm{cond}} \to 4d_{\mathrm{cond}} \to d_{\mathrm{model}}$,
  GELU nonlinearity), and added to every token position (broadcast  over the sequence).
  \item \textbf{Backbone.} $n_{\mathrm{layers}}$ pre-norm
  blocks ($d_{\mathrm{model}}$,
  $n_{\mathrm{heads}}$ heads, feed-forward width
  $4d_{\mathrm{model}}$, GELU, dropout $p$), full self-attention.
  \item \textbf{Output head.} LayerNorm $\to$ linear projection to $3$
  logits, followed by the SUBS parameterization of MDLM: the MASK logit is set to $-\infty$; at
  unmasked positions the output distribution is forced to a point
  mass on the observed token; at masked positions the model outputs a
  free $\mathrm{softmax}$ over $\{0,1\}$. The output projection is
  zero-initialized.
\end{itemize}
For the run analyzed here: $d_{\mathrm{model}}=64$,
$n_{\mathrm{heads}}=4$, $n_{\mathrm{layers}}=2$,
$d_{\mathrm{cond}}=64$, dropout $p=0.1$, giving $\approx 1.85\times
10^{5}$ trainable parameters.

Training parameters are reported below

\begin{table}[h]
\centering
\begin{tabular}{ll}
\toprule
Optimizer & AdamW, weight decay $10^{-2}$ \\
Peak learning rate & $3\times10^{-2}$ \\
LR schedule & linear warm-up ($50$ steps) $\to$ constant $\to$
  linear decay to $10^{-3}\times$ peak over the \\
  & final $20\%$ of training (decay starts at step $80{,}000$) \\
Batch size & $256$ \\
Training steps & $100{,}000$ \\
Gradient clipping & global norm $1.0$ \\
Seeds & $4$ independent runs \\
\bottomrule
\end{tabular}
\caption{Optimization hyperparameters for the run shown in Fig. \ref{fig:mdlm-humgen}}
\end{table}

\subsubsection{Figure \ref{fig:mdlm-humgen}: PCA-learning collapse }
\label{sec:metrics-genome}

At every logged training step $s$, and for each of the $5{,}000$
freshly generated samples, we:
\begin{enumerate}
  \item Fit a fresh $K$-component PCA, $\{\vmu_k^{\,\mathrm{inf}}(s)\}$,
  on the generated batch (``inferred'' PCA at step $s$).
  \item Compute the overlap matrix between the inferred and the fixed
  reference eigenvectors,
  \begin{equation}
    O_{ij}(s) = \frac{1}{|\vmu_j^{\rm ref}|^2}\,\vmu_i^{\,\mathrm{inf}}(s) \cdot \vmu_j^{\rm ref},
    \qquad i,j = 1,\dots,K,
  \end{equation}
  \item Pair inferred and reference directions one to one, through the
  permutation $\pi_s$ of $\{1,\dots,K\}$ maximizing the total matched overlap
  $\sum_j|O_{\pi_s(j)j}(s)|$ (Hungarian algorithm). The matched value
  $|\vmu_{\mathrm{inf}}\!\cdot\!\vmu_{\mathrm{ref}}|_k(s):=|O_{\pi_s(k)k}(s)|$
  tracks how well the $k$-th reference principal direction has emerged in the generated samples by training step $s$.
\end{enumerate}
Ranking the entries of $O$ and reading the $k$-th largest as the $k$-th direction
would be ambiguous on two counts: the curves would then be non-increasing in $k$
by construction, whatever the model has learned, and a single dominant inferred
component leaking onto several reference eigenvectors could occupy several of the
top slots. Both matter here, since the inset of Fig.~\ref{fig:mdlm-humgen}
rescales curve $k$ by the timescale of reference direction $k$ specifically. The
two constructions agree on the two leading directions, each carried by a single
inferred component, and differ on the subleading ones.
The reported curves are the mean and $95\%$ CI of
$|\vmu_{\mathrm{inf}}\!\cdot\!\vmu_{\mathrm{ref}}|_k(s)$ across the $4$
training seeds, lightly smoothed along the
step axis (Gaussian filter, $\sigma=0.1$ index units), for the top
$5$ modes.

\paragraph*{Left panel.} A plain scatter of $\vx^{T}\vmu_0$ vs.\ $\vx^{T}\vmu_1$
(the projections of each haplotype onto the top-two reference PCA
directions), using every one of the $N_{\mathrm{train}}=4{,}508$ real
training haplotypes -- no subsampling. Both the scatter and the
reference directions $\{\vmu_k\}$ come from the same reference PCA fit.

\paragraph*{Right panel.}
$|\vmu_{\mathrm{inf}}\!\cdot\!\vmu_{\mathrm{ref}}|_k$, labeled \textbf{PCA Overlap} on the figure, plotted against the raw training step $s$; the inset shows the same curves against the rescaled step $\lambda_k\times s$, where they collapse.
To identify the rescaling timescales we extract the explained variance $v_k$ along each of the reference PCA directions. The GMM growth rate \eqref{eq:mu_main} is a function of the block SNR $\gamma_i$, which at the speciation scale reduces to $\gamma_i=\sqrt{\kappa_i}$, with $\kappa_i=\lVert\vmu_i\rVert^2/d$ the squared amplitude of direction $i$. Identifying that squared amplitude with the variance explained along direction $k$, we rescale the time axis by
\begin{equation}
    \lambda_k \equiv \lambda\left(\sqrt{v_k}\right).
\end{equation}
The PCA projected Human Genome dataset is of course not a GMM, and the rescaling should only be expected to hold where the projected data actually resolves into modes. This is the case for the two leading directions, which collapse onto one another; the subsequent ones, along which no mode separation is visible, do not, and the ansatz carries less meaning there.

\section{LLM usage}

We acknowledge the use of LLMs to assist in drafting the manuscript and refining its clarity. The tool was also used to support the derivation of mathematical proofs and the writing of the code. All AI-generated content was rigorously reviewed, verified, and edited by the authors. The authors bear full responsibility for the originality and scientific integrity of this work. \looseness=-1

\end{document}